\documentclass[12pt]{article}
\usepackage{amsmath,amssymb}
\usepackage{amsthm} 
\usepackage{graphicx}
\usepackage{multirow}
\usepackage[unicode=true,backref=page]{hyperref}
\usepackage{algorithm}
\usepackage{algpseudocode}
\usepackage{subcaption}
\usepackage[authoryear]{natbib}
\usepackage{rotating}
\usepackage{graphicx}
\usepackage{array}
\usepackage{setspace}
\usepackage[authoryear]{natbib}
\usepackage{bbm}
\usepackage{titlesec}
\titleformat*{\section}{\normalsize\bfseries}
\titleformat*{\subsection}{\normalsize\bfseries}

\newcolumntype{P}[1]{>{\centering\arraybackslash}p{#1}}

\newcommand{\bb}[1]{\boldsymbol{#1}}
\newcommand{\minimize}{\operatorname*{minimize}}
\newcommand{\argmin}{\operatorname*{arg\,min}}

\newcommand{\err}{\operatorname{err}}

\theoremstyle{definition}
\newtheorem{definition}{Definition}
\newtheorem{theorem}{Theorem}
\newtheorem{lemma}{Lemma}
\newtheorem{theoremA}{Theorem}

\algrenewcommand\algorithmicrequire{\textbf{Input:}}
\algrenewcommand\algorithmicensure{\textbf{Output:}}

\begin{document}
\hspace{13.9cm}

\ \vspace{20mm}\\

{\LARGE \noindent Unsupervised Domain Adaptation for\\ Extra Features in the Target Domain\\Using Optimal Transport}

\ \\
{\bf \large Toshimitsu Aritake$^{\displaystyle 1}$, Hideitsu Hino$^{\displaystyle 1, \displaystyle 2}$}\\
{$^{\displaystyle 1}$The Institute of Statistical Mathematics, Tachikawa, Tokyo, Japan}\\
{$^{\displaystyle 2}$RIKEN AIP, Tokyo, Japan}\\


\ \vspace{-0mm}\\
\begin{center} {\bf Abstract} \end{center}
Domain adaptation aims to transfer knowledge of labeled instances obtained from a source domain to a target domain to fill the gap between the domains. Most domain adaptation methods assume that the source and target domains have the same dimensionality. Methods that are applicable when the number of features is different in each domain have rarely been studied, especially when no label information is given for the test data obtained from the target domain. In this paper, it is assumed that common features exist in both domains and that extra (new additional) features are observed in the target domain; hence, the dimensionality of the target domain is higher than that of the source domain.
To leverage the homogeneity of the common features, the adaptation between these source and target domains is formulated as an optimal transport (OT) problem.
In addition, a learning bound in the target domain for the proposed OT-based method is derived.
The proposed algorithm is validated using both simulated and real-world data.

\section{Introduction}
The goal of supervised learning is to build a model $f$ that maps the feature $\bb x$ to its corresponding label $y$ from a given training dataset $\mathcal{D}_S$ to estimate the label of the unlabeled test dataset $\mathcal{D}_T$.
Let the distributions of the training data be $\mathcal{P}_S(\bb x, y)$, and the test data be $\mathcal{P}_T(\bb x, y)$, respectively.
In a supervised learning framework, it is generally assumed that the training and test data follow the same distribution. 
However, when $\mathcal{P}_S(\bb x, y)\neq \mathcal{P}_T(\bb x, y)$, the difference leads loss of accuracy of the trained model on the test data.
It is still possible to train a model that accurately predicts the label of the test data by considering the difference in the distributions of the training and test data. 
Domain adaptation techniques are used to consider the difference in the distributions by transferring information from the source domain to the target domain~\citep{Ben-David2010,redko2019advances}. 
Henceforth, we refer to the domains of the training and test data as the source and target domains, respectively.
In general, domain adaptation aims to match the joint distributions of $(\bb{x},y)$ in the source and target domains.

In this paper, we consider an unsupervised domain adaptation problem, where both the source and target domains have common features and extra (new additional) features are observed in the target domain.
Here, we assume that data is tabular data and that it is known whether each feature is a common feature or an extra feature.
Also, since each common feature represents the same attribute in the source and target domains, the homogeneity of common features should be considered for domain adaptation.

For example, consider the case of measuring the movements of a person with a set of accelerometers. 
The types of activity are assigned as a label for each observed movement, and these data are used for training.
Then, assume that the activities of another person are estimated from the measurements of movements obtained using the same set of accelerometers and additional gyroscopes.
In this case, the features obtained by the accelerometers become common features, and the features obtained by the gyroscopes become extra features.

However, most domain adaptation methods assume spaces of the same dimensionality as the source and target domains.
This type of domain adaptation is called \textit{homogeneous} domain adaptation, and these methods cannot be applied when the number of features is different for each domain.
Domain adaptation for spaces of different dimensionalities is called \textit{heterogeneous} domain adaptation.
In the literature, only a few methods have been proposed for unsupervised heterogeneous domain adaptation.
Furthermore, general heterogeneous domain adaptation methods cannot consider the homogeneity of the common features between the source and target domains.

To address this issue, a special case of heterogeneous domain adaptation called hybrid domain adaptation has been studied, where it is assumed that the source and target domains have common features, and domain-specific features are also given for each domain.
To preserve the homogeneity of the common features, hybrid domain adaptation learns the models used to predict domain-specific features from common features.
Then, the learned models are used to estimate the unobserved domain-specific features.

The problem considered in this paper can be seen as a variant of a hybrid domain adaptation problem, in which the domain-specific features are only given for the target domain.
In the same manner as for hybrid domain adaptation, the unobserved extra features in the source domain are predicted using common features.
Unlike hybrid domain adaptation, which learns a model used to estimate the unobserved features, our proposed method estimates the unobserved features using optimal transport (OT).
Recently, OT has been used for domain adaptation to match the distributions in the source and target domains.
However, when the number of features is different between the source and target domains, it is difficult to define an appropriate transport cost for OT.

\begin{figure}[b!]
    \centering
    \includegraphics[width=.60\textwidth]{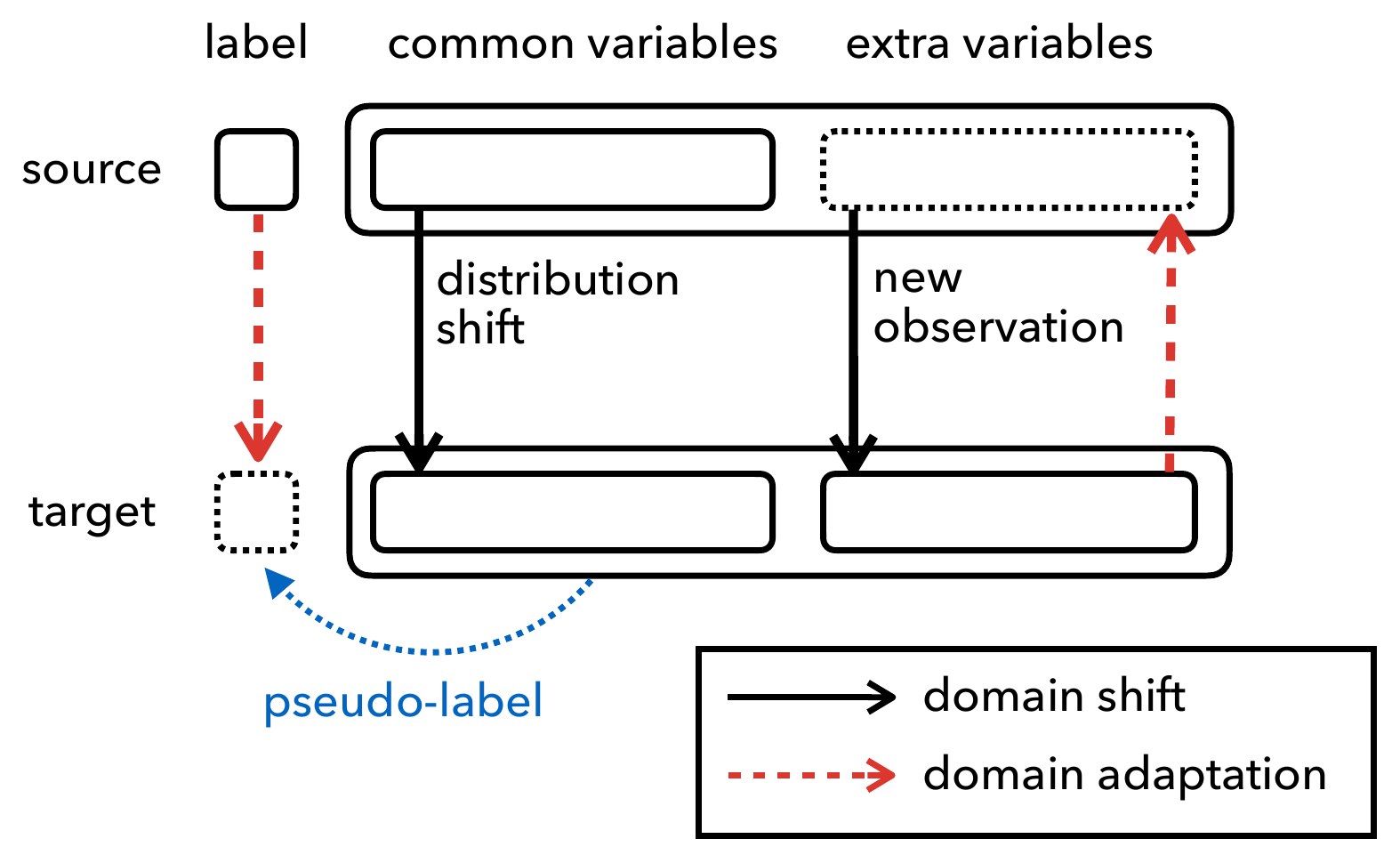}
    \caption{Conceptual illustration of the proposed domain adaptation method for the observation of extra features. The two-way OT between the source and target domains is considered to estimate the extra features in the source domain and the labels in the target domain. Practically, this two-way OT is solved as one-way OT from the source domain to the target domain.}
    \label{fig:conceptual_illustration}
\end{figure}

To solve our problem using OT, it is natural to consider \textit{two-way} OT.
Namely, extra features in the source domain are estimated by solving the OT problem from the target domain to the source domain, then the label information in the source domain is transferred to the target domain by solving another OT problem.
For these OT problems, we use pseudo-labels as proxies of unobserved true target labels and consider a problem similar to joint distribution optimal transport (JDOT)~\citep{NIPS2017_0070d23b}.
Namely, in the former OT, the distance between the common features and the mismatch between the source label and the target pseudo-label are used for the transport cost so that the joint distributions of the features and labels are better matched in the source and target domains.
Then, in the latter OT, the distance between the extra features is additionally considered.
We show that this \textit{two-way} OT is equivalent to \textit{one-way} OT under the assumption that the conditional distribution of an extra feature given a common feature and a label is identical before and after OT.
Figure~\ref{fig:conceptual_illustration} shows the above concept.

We summarize the contributions of this paper:
\begin{enumerate}
     \item We propose an algorithm based on OT for a domain adaptation problem where the domain shift between the source and target domains is caused by the observation of extra (new additional) features and the distribution shift of common features.
    \item We provide an interpretation of the proposed algorithm that the proposed one-way OT-based algorithm is equivalent to two-way OT.
    \item We derive a learning bound of the trained model by the proposed method in the target domain. The derived upper bound is based on the Rademacher complexity and the Wasserstein distance between the true and estimated target distributions.
    The upper bound using only a Wasserstein distance between the source and target distributions is already proposed in the previous work.
    We refined the upper bound analysis by using Rademacher complexity to consider the training of a model in the target domain.
    The upper bound also gives an intuitive understanding of the proposed algorithm.
\end{enumerate}

The rest of this paper is organized as follows.
In Section \ref{sec:related_work}, the related work of domain adaptation and OT is summarized.
In Section \ref{sec:problem_formulation}, we present the practical algorithm based on one-way OT.
Then, we show the equivalence of the proposed method and two-way OT in Section \ref{sec:theory}.
Also, the learning bound of the proposed method is presented.
In Section \ref{sec:experiments}, we report the results of experiments on synthetic and real-world datasets.
Then, we summarize the paper and discuss the limitations of the proposed method and future work in Section \ref{sec:conclusion}.

\section{Related Work}
\label{sec:related_work}

\subsection{Domain Adaptation}
In general, domain adaptation aims to match the joint distributions of the features and the label $(\bb{x},y)$ in the source and target domains.
When it is possible to access the label information of the target domain, the problem of domain adaptation is categorized into \textit{(semi-)supervised} domain adaptation depending on the accessibility of the label information.
On the other hand, when no labeled data of the target domain are available, the domain adaptation problem is called \textit{unsupervised} domain adaptation, which we consider in this paper.

Most domain adaptation methods assume spaces of the same dimensionality as the source and target domains.
This type of domain adaptation problem is called \textit{homogeneous} domain adaptation.
On the other hand, when the source and target domains have different dimensionalities, the domain adaptation problem is called \textit{heterogeneous} domain adaptation.
\subsubsection{Homogeneous Domain Adaptation}
In unsupervised homogeneous domain adaptation, the distributions of the source and target domains are matched on the basis of the assumption made for the change in distribution.
There are a number of unsupervised domain adaptation methods, which are categorized into several groups.
The covariate shift~\citep{Shimodaira2000} assumes that $\mathcal{P}_S(y|\bb x) = \mathcal{P}_T(y|\bb x)$ and $\mathcal{P}_S(\bb x)\neq \mathcal{P}_T(\bb x)$.
Therefore, it aims to match the distributions $\mathcal{P}_S(\bb x)$ and $\mathcal{P}_T(\bb x)$ to match the joint distribution, for example, by importance reweighting~\citep{10.5555/2981562.2981742}.
Similarly, the conditional shift or the concept shift~\citep{10.1023/A:1018046501280} assumes either $\mathcal{P}_S(y|\bb x) \neq \mathcal{P}_T(y|\bb x)$ or $\mathcal{P}_S(\bb x|y) \neq \mathcal{P}_T(\bb x|y)$, and the target shift (also known as the prior shift)~\citep{Webb2005,10.1007/978-3-540-68825-9_2} assumes $\mathcal{P}_S(y)\neq \mathcal{P}_T(y)$.
Furthermore, recent works have considered to learn domain invariant features for each of these assumptions using deep neural networks including generative adversarial models~\cite{8099799,aaai/ShenQZY18,ZHOU2021106986,pmlr-v97-zhao19a}.
For other domain adaptation methods, eee~\citep{MORENOTORRES2012521,PMID:31603771} and references therein.

These homogeneous domain adaptation methods rely on the assumption that the source and target domains have the same dimensionality; therefore, these methods are not directly applicable when the source and target domains have different dimensionalities.
\subsubsection{Heterogeneous Domain Adaptation}
Heterogeneous domain adaptation is considered for the domain adaptation between the domains that have different dimensionalities.
In the literature, several methods have been proposed to solve heterogeneous domain adaptation problems.
However, most heterogeneous domain adaptation methods~\citep{6104043,Wang2011,Li2014,6866177} require at least partly labeled instances from the target domain, and only a few unsupervised heterogeneous domain adaptation methods have been proposed~\citep{Yeh2014,8432087,Zhou_Pan_Tsang_Yan_2014}.

The common strategy for unsupervised heterogeneous domain adaptation is to embed features from the source and target domains to a space of the same dimensionality and consider a homogeneous domain adaptation problem therein.
For example, spectral embedding~\citep{6104043}, linear embedding~\citep{Zhou_Pan_Tsang_Yan_2014}, and kernel canonical correlation analysis~\citep{Yeh2014} are used for embedding. 
However, since the features are mixed by embedding, these methods cannot consider the homogeneity of features even when the source and target domains have common features.

\subsubsection{Hybrid Domain Adaptation}
A special case of heterogeneous domain adaptation called hybrid domain adaptation is studied in~\citep{8432087,Prabono2021}, where it is assumed that the source and target domains have common features, and domain-specific features are also given for each domain.
To consider the homogeneity of the common features, hybrid domain adaptation use the models to predict domain-specific features from common features.
The models are learned on one domain, and the learned models are used to estimate the unobserved domain-specific features on the other domain.
However, to consider the shift of common features, the embedding of common features are used in~\citep{Prabono2021}, and original common features are mixed with other common features.
Therefore, although the dimensionality of common features is preserved, embedded features are not homogeneous due to the mixture of the features.
In addition, it is not always possible to accurately estimate the domain-specific features from the common features.
For example, it is difficult to estimate domain-specific features using simple regression models when the distribution of domain-specific features given the common features follow multi-modal distributions.

\subsection{Optimal Transport in Domain Adaptation}
Recent works apply OT techniques to match the source and target distributions for domain adaptation~\citep{ijcai2020-299,NIPS2017_0070d23b,7586038}.
The OT problem is a well-established mathematical theory~\citep{villani2008optimal}, which has been successfully applied to various machine learning tasks that involve the transport of a probability distribution.
The OT for homogeneous domain adaptation makes the assumption $\mathcal{P}_S(y|\bb x) = \mathcal{P}_T(y|\mathcal{T}(\bb x))$ on the conditional distribution, where $\mathcal{T}$ represents the OT.
However, this assumption does not hold in general; therefore, group regularized OT~\citep{7586038} and JDOT~\citep{NIPS2017_0070d23b}, which leverages pseudo-labels estimated using the model, are proposed to alleviate this problem.
In addition, some works have considered OT problems for heterogeneous feature spaces by defining the transport cost between spaces of different dimensionalities~\citep{pmlr-v48-peyre16,NEURIPS2020_cc384c68}.
Although these methods are applicable for heterogeneous domain adaptation, the cost functions defined in these methods do not consider the homogeneity of features.

Our proposed method also use OT for domain adaptation, and is categorized into these groups of works.
A short and preliminary version of this paper appeared in the 2022 International Joint Conference on Neural Networks (IJCNN 2022)~\citep{wcci2022_OTDA}.

\section{Problem Formulation}
\label{sec:problem_formulation}
\subsection{Optimal Transport in Domain Adaptation}
Let $\mu_1$ and $\mu_2$ be the probability measure on a space $\Omega$.
Given a cost function $c:\Omega\times\Omega\to\mathbb{R}_+$, the problem of OT is formulated as a problem of seeking a coupling $\pi\in\Pi(\mu_1, \mu_2)$ between $\mu_1$ and $\mu_2$ that minimizes the total transport cost:
\begin{equation}
\inf_{\pi\in\Pi(\mu_1,\mu_2)}\int_{\Omega\times \Omega} c(\bb x, \bb x') d\pi(\mu_1, \mu_2),
\end{equation}
where $\bb x\sim\mu_1$ and $\bb x'\sim\mu_2$.
Here, $\Pi(\mu_1, \mu_2)$ is the set of couplings, that is, joint probability distributions with marginals $\mu_1$ and $\mu_2$.
In general, a distance function between the samples is used as a cost function $c$.

In practice, a discretized version of the above problem is solved.
Here, we define two datasets, $\mathcal{D}_1=\{\bb x_i\mid \bb x_i\sim \mu_1\ (i=1,2,\ldots ,n_1)\}$ and $\mathcal{D}_2=\{\bb x'_j \mid \bb x'_j\sim \mu_2\ (i=1,2,\ldots ,n_2)\}$, and let $\nu_1$ and $\nu_2$ be the empirical distributions of $\mathcal{D}_1$ and $\mathcal{D}_2$, respectively.
Then, OT between $\nu_1$ and $\nu_2$ is formulated as
\begin{equation}
\minimize_{\pi\in\Pi(\nu_1,\nu_2)}\sum_{i=1}^{n_1}\sum_{j=1}^{n_2} \pi_{ij}c(\bb x_i, \bb x'_j).
\label{eq:discrete_ot_general}
\end{equation}
Here, both coupling and transport cost can be treated as $n_1\times n_2$ matrices; hence, the above problem can be solved as a linear programming problem.

In domain adaptation, OT is used to match the source distribution $\mathcal{P}_S(\bb x, y)$ and the target distribution $\mathcal{P}_T(\bb x, y)$.
In particular, the OT problem for \textit{unsupervised} domain adaptation is formulated as OT between two marginal distributions $\mathcal{P}_S(\bb x)$ and $\mathcal{P}_T(\bb x)$ under the assumption $\mathcal{P}_S(y|\bb x) = \mathcal{P}_T(y|\mathcal{T}(\bb x))$, where $\mathcal{T}$ represents an optimal transport map.
However, this assumption does not always hold; hence, JDOT~\citep{NIPS2017_0070d23b} considers the distance between features as well as discrepancy of the labels as transport cost so that $\mathcal{P}_S(\bb x, y)$ and $\mathcal{P}_T(\bb x, y)$ are better matched.
Namely, the cost function $c(\bb x_1, y_1; \bb x_2, y_2)=\alpha d(\bb x_1, \bb x_2) + \mathcal{L}(y_1, y_2)$, where $\mathcal{L}$ is the discrepancy between labels $y_1$ and $y_2$, is used for OT.
Since the target label is not observed in unsupervised domain adaptation, the label estimated as $\hat{y}=f(\bb x)$ is used as a proxy of the target label.
The technique of using the output of the model as the proxy of the true label is known as pseudo-labeling, and a number of methods that use pseudo-labeling have been proposed for various learning problems including unsupervised domain adaptation~\citep{lee2013pseudo,NIPS2017_0070d23b,Damodaran_2018_ECCV,Zou_2018_ECCV,DBLP:conf/eccv/ShinWPK20,DBLP:conf/icml/XieZCC18,DBLP:conf/cvpr/ZhangO0018,DBLP:conf/cvpr/Kang0YH19}.

Furthermore, JDOT learns a model $f$ that estimates the pseudo-label in the target domain.
In summary, the problem of JDOT can be written as
\begin{align}
        &\inf_{\substack{\pi\in\Pi(\mathcal{P}_S,\mathcal{P}_T^f)\\f\in\mathcal{F}}}\int_{(\Omega\times\mathcal{C})^2} c(\bb x_1, y_1; \bb x_2, y_2)d\pi(\bb x_1, y_1; \bb x_2, y_2),
\end{align}
where $\mathcal{F}$ is a set of models, $\mathcal{C}$ is a label space, and $\mathcal{P}_T^f=(\bb x, f(\bb x))_{\bb x\sim\mathcal{P}_T(\bb x)}$.
In practice, the discretized version of the above problem is solved as Eq.~\eqref{eq:discrete_ot_general}.

\subsection{Domain Adaptation with Optimal Transport for Extended Feature Space}
Let $\Omega^c \times \mathcal{C}$ be the source domain, which is a direct product of the space of the common features, $\Omega^c$, and label space $\mathcal{C}$.
Also, we define $\Omega^t\equiv \Omega^c\times\Omega^e$, where $\Omega^e$ is the space of the extra features, and let $\Omega^t \times \mathcal{C}$ be the target domain. 
Namely, the spaces of the common features are identical in the source and target domains, and the extra features are only observed in the target domain. 
Note that even though the spaces of the common features are identical, the distribution on $\Omega^c$ can be different between the source and target domains. 
Here, we denote the probability distributions of the source and target domains as $\mathcal{P}_S(\bb x^c, y)$, and $\mathcal{P}_T(\bb x^c, \bb x^e, y)$, respectively, or $\mathcal{P}_S$ and $\mathcal{P}_T$ for short.
Then, we define the training set $\mathcal{D}_S = \{(\bb x^c_{si}, y_{si})\}_{i=1}^{N_s}$ that consists of samples $(\bb x^c_{si}, y_{si})\sim \mathcal{P}_S$. Similarly, we define the test set by the partial observation  $\mathcal{D}_T = \{(\bb x^c_{ti}, \bb x^e_{t,i})\}_{i=1}^{N_t}$ of a sample $(\bb x^c_{ti}, \bb x^e_{ti}, y_{ti}) \sim\mathcal{P}_T\ (i=1,\dots,N_t)$, where the true label $y_{ti}$ is not observed.
Since the label $y$ of the target distribution is not observed, we define the estimated target probability distribution as $\mathcal{P}_T^f(\bb x^c, \bb x^e, \hat{y})$, where the label $y$ is replaced by the pseudo-label $\hat{y}=f(\bb x^c,\bb x^e)$.
Note here that $\mathcal{P}_T^f(\bb x^c, \bb x^e)=\mathcal{P}_T(\bb x^c, \bb x^e)$ holds for marginal distributions.
Similarly, we define the estimated test set $\mathcal{D}_T^f=\{(\bb x^c_{ti}, \bb x^e_{ti}, \hat{y}_i\}_{i=1}^{N_t}$, where $\hat{y}_i = f(\bb x^c_{ti}, \bb x^e_{ti})$.

To transfer the label information from the source domain to the target domain,
we consider the following problem, which is similar to the problem of JDOT:
\begin{align}
\pi^*, \hat{f} = \inf_{\substack{\pi\in\Pi(\mathcal{P}_S, \mathcal{P}_T^f)\\f\in\mathcal{F}}}\int_{\substack{(\Omega_s\times \mathcal{C})\\\times (\Omega_t\times\mathcal{C})}}\mathcal{E}_\alpha(\bb x^c_s, y_s; \bb x^c_t, \bb x^e_t, y_t)d\pi(\bb x^c_s, y_s;\bb x^c_t, \bb x^e_t, y_t),\label{eq:optimal_transport}
\end{align}
where $\Pi(\mathcal{P}_S,\mathcal{P}_T^f)$ is the set of transportation plans between the probability densities $\mathcal{P}_S$ and $\mathcal{P}_T^f$ and $\mathcal{F}$ is a set of models.
Here, we use the following cost function for the transport:
\begin{equation}
\mathcal{E}_\alpha(\bb x^c_s,y_s;\bb x^c_t, \bb x^e_t, y_t)\equiv \alpha d(\bb x^c_s, \bb x^c_t) + \mathcal{L}(y_s, y_t),
\label{eq:transportation_cost}
\end{equation}
which is the sum of the distance between the common features $d(\bb x_s^c, \bb x_t^c)$ and the discrepancy of the label $\mathcal{L}(y_s, y_t)$.
Note here that the extra feature $\bb x^e_t$ is only used to estimate the pseudo-label $\hat{y}=f(\bb x^c,\bb x^e)$.
Although the choice of the metric $d$ is arbitrary, here we assume that $d$ is a square distance $d(\bb x_s^c, \bb x_t^c)= \|\bb x_s^c-\bb x_t^c\|_2^2$ for simplicity of description.
Here, $\alpha\in\mathbb{R}_+$ is a hyperparameter that determines the relative importance of $d(\bb x_s^c,  \bb x_t^c)$ to $\mathcal{L}(y_s, y_t)$.
By solving the above optimization problem, the source labels are transferred to the target domain, and model $f\in\mathcal{F}$ is trained to map the pair of common and extra features to their corresponding transferred labels.

In practice, a finite number of samples obtained from the source and target distributions can be used to solve the OT problem. Therefore, instead of solving the OT problem between the source and target distributions, we consider the discrete OT problem between the empirical distributions of the training and test data.
The optimization problem Eq.~\eqref{eq:optimal_transport} is rewritten as
\begin{equation*}
    \hat{\pi}^*, \hat{f}_s 
    =\argmin_{\hat{\pi}\in\hat{\Pi}(\mathcal{D}_S,\mathcal{D}_T^f),f\in\mathcal{F}}\sum_{i=1}^{N_s}\sum_{j=1}^{N_t} \hat{\pi}_{ij}\mathcal{E}_\alpha(\bb x^c_{si},y_{si};\bb x^c_{tj}, \hat{y}_{tj}),
\end{equation*}
where $\hat{y}_{tj}=f(\bb x^c_{tj},\bb x^e_{tj})$ and $\hat{\Pi}(\mathcal{D}_S,\mathcal{D}_T^f)$ is a set of discrete OT plans from the dataset $\mathcal{D}_S$ to the dataset $\mathcal{D}_T^f$, and is defined as
\begin{align*}
    &\hat{\Pi}(\mathcal{D}_S,\mathcal{D}_T^f)\equiv \left\{\pi\in\mathbb{R}_+^{N_s\times N_t}\middle|\textstyle\sum_{i=1}^{N_s}\pi_{ij}=\frac{1}{N_t}, \sum_{j=1}^{N_t}\pi_{ij}=\frac{1}{N_s}\right\}.
\end{align*}

This optimization problem is non-convex and computationally intractable; therefore, alternating optimization is used to solve the problem in the same manner as in conventional methods that use pseudo-labeling, such as in~\citep{lee2013pseudo,NIPS2017_0070d23b,Damodaran_2018_ECCV,Zou_2018_ECCV,DBLP:conf/eccv/ShinWPK20}.
At the $n$th iteration, the optimization problem with respect to $\pi\in\hat{\Pi}(\mathcal{D}_S, \mathcal{D}_T^f)$ with fixed $\hat{f}_s^{(n)}\in\mathcal{F}$ becomes a discrete OT problem, where the transport cost $\mathcal{E}_\alpha$ is calculated using the pseudo-label estimated as $\hat{y}^{(n)}=\hat{f}_s^{(n)}(\bb x^c_t, \bb x^e_t)$.
Here, the optimization problem is
\begin{align}
    \hat{\pi}^*_n=&\argmin_{\hat{\pi}\in\hat{\Pi}(\mathcal{D}_S,\mathcal{D}_T^f)}
    \sum_{i=1}^{N_s}\sum_{j=1}^{N_t} \hat{\pi}_{ij}\mathcal{E}_\alpha(\bb x^c_{si},y_{si};\bb x^c_{tj}, \hat{y}^{(n)}).
    \label{eq:discret_ot}
\end{align}
This problem can be solved as a linear programming problem.
The model $\hat{f}_s^{(1)}$ is not obtained for the first iteration; hence, the cost $\mathcal{E}^0(\bb x^c_s;\bb x^c_t, \bb x^e_t)\equiv d(\bb x_{si}^c, \bb x_{tj}^c)$ is used instead of $\mathcal{E}_\alpha$ only for the first iteration.

Then, the optimization problem with respect to $f$ with a fixed $n$th OT plan $\hat{\pi}^*_n$ is solved to train model $f$, namely,
\begin{equation}
\hat{f}_s^{(n+1)} = \argmin_{f\in\mathcal{F}}\sum_{i=1}^{N_s}\sum_{j=1}^{N_t} (\hat{\pi}^*_n)_{ij}\mathcal{L}(y_{si}, f(\bb x^c_{tj}, \bb x^e_{tj})).
    \label{eq:update_f}
\end{equation}
In the above problem, there are cases in which different labels are transferred onto one sample.
In these cases, we use the weighted average of the source labels as transferred labels.
Then, the transferred labels can be calculated by barycentric mapping using $\hat{\pi}^*_n$.
When the distance between the common features is the squared distance, we obtain
\begin{equation*}
    \tilde{\bb y}_t = \operatorname{diag}(\bb 1^\top \hat{\pi}^*_n)^{-1}(\hat{\pi}^*_n)^\top \bb y_s.
\end{equation*}
In particular, when the task is classification, the assigned labels can be seen as soft class labels.
Here, let $\bb Y_s$ be the label matrix whose rows are one-hot encoded labels of source samples, namely,
\begin{equation*}
    Y_{s,ik} = \begin{cases}
    1& y_i = k\in\mathcal{C},\\
    0 & \text{otherwise}.
    \end{cases}
\end{equation*}
Then the transferred soft-labels are calculated by $\tilde{\bb Y}_t = \operatorname{diag}(\bb 1^\top \hat{\pi}^*_n)^{-1}(\hat{\pi}^*_n)^\top \bb Y_s$, and rows of transferred label matrix $\tilde{\bb Y}_t$ can be seen as class probabilities.
However, to train model $f$, it would be easier to use hard labels, then we use hard labels
\begin{equation}
    \bar{y}_{tj} =
    \operatorname*{arg\,max}_{k} \tilde{Y}_{t,jk}.
    \label{eq:transport_label}
\end{equation}
We remark that if the $j$th row has equal probabilities for all classes, we assign a random label that follows uniform distribution.
Then, the training of the model Eq.~\eqref{eq:update_f} becomes the following simple training process in the target domain:
\begin{equation}
\hat{f}_s^{(n)} = \argmin_{f\in\mathcal{F}}\sum_{j=1}^{N_t}\mathcal{L}(\bar{y}_{tj}, f(\bb x^c_{tj}, \bb x^e_{tj})).
    \label{eq:update_f2}
\end{equation}
The above algorithm using the hard labels is summarized as Algorithm~\ref{alg:alg1}.
\begin{algorithm}[ht]
\caption{Domain adaptation with extra features in target domain}\label{alg:alg1}
\begin{algorithmic}
\Require datasets $\mathcal{D}_S, \mathcal{D}_T$, model set $\mathcal{F}$, number of iterations $N$
\Ensure Optimal transport plan $\hat{\pi}^*_N$, trained model $\hat{f}_{s}^{(N)}$

\State $n \gets 1$
\While{$n < N$}
\If{$n = 1$}
\State $ \displaystyle\hat{\pi}^*_1 \leftarrow\argmin_{\pi\in\hat{\Pi}(\mathcal{D}_S,\mathcal{D}_T)}\sum_{i=1}^{N_s}\sum_{j=1}^{N_t} \pi_{ij}\mathcal{E}^0(\bb x^c_{si}, \bb x^c_{tj})$
\Else
\State $ \displaystyle\hat{\pi}^*_n \leftarrow \argmin_{\pi\in\hat{\Pi}(\mathcal{D}_S,\mathcal{D}_T)}\sum_{i=1}^{N_s}\sum_{j=1}^{N_t}$\\\hspace{7em}$\pi_{ij}\mathcal{E}(\bb x^c_{si}, y_{si}; \bb x^c_{tj}, \hat{f}_{s}^{(n)}(\bb x^c_{tj}, \bb x^e_{tj}))$
\EndIf
\State $\tilde{\bb Y}_t \leftarrow \operatorname{diag}(\bb 1^\top \hat{\pi}^*_n)^{-1}(\hat{\pi}^*_n)^\top \bb Y_s$
\State estimate hard labels $\bar{y}_{tj}\ (j=1,\ldots N_t)$ by Eq.~\eqref{eq:transport_label}
\State $\hat{f}_{s}^{(n+1)}\leftarrow \argmin_{f\in\mathcal{F}}\sum_{j=1}^{N_t}\mathcal{L}(\bar{y}_{tj}, f(\bb x^c_{tj}, \bb x^e_{tj}))$
\State $n\gets n+1$
\EndWhile
\end{algorithmic}
\end{algorithm}

\section{Theoretical Justification And Analysis of Proposed Algorithm}
\label{sec:theory}
In this section, the proposed method is analyzed mainly from two perspectives.
First, we give an interpretation of our proposed method.
Briefly, the main optimization problem Eq.~\eqref{eq:optimal_transport} of the proposed method is identical to the two-way OT between the source and target domains under an assumption that the conditional distributions of $\bb x^e$ given $\bb x^c$ and $y$ in the source and target domains are identical.

Then, a learning bound of the model $f$ on the target domain is derived.
The conventional analyses of domain adaptation methods based on OT~\citep{redko:hal-01613564,NIPS2017_0070d23b} give learning bounds that mainly focus on the Wasserstein distance between the source and target distributions.
Although it is possible to extend this upper bound for our algorithm, they become loose when the Wasserstein distance between the source and target distributions becomes large even if it is possible to correctly transfer source labels to the target domain.
Moreover, the upper bound does not consider how the model $f$ is trained in the target domain.
On the other hand, we gives an upper bound that focuses on the training of the model $f$ in this paper, and the target error is upper bounded by the Rademacher complexity and the Wasserstein distance between the estimated and true target distributions.
By using Rademacher complexity, we can include the empirical distribution of the estimated target distribution, which is actually used for training a model in the target domain, into the derived upper bound.
That is, the upper bound becomes tight when the transferred source distribution is close to the true target distribution, and the model can accurately predict the transferred label.
This interpretation gives an intuitive understanding of the condition required for the successful domain adaptation.

\subsection{Theoretical Justification of Proposed Algorithm}
The main problem stated in Eq.~\eqref{eq:optimal_transport} considers the transportation from the source domain to the target domain.
Let us start with an ideal case that the common feature $\bb x^c$, the extra feature $\bb x^e$, and the label $y$ are observed in both the source and target domains.
Here, let $\bb x^e_s$ be the extra features in the source domain, which are not observed in practice.
In this ideal case, the domain adaptation becomes a homogeneous domain adaptation problem, which is relatively easy to solve.
The cost function for the transportation is defined as
\begin{align}
    \mathcal{E}_\alpha^*(\bb x^c_s, \bb x^e_s, y_s; \bb x^c_t,\bb x^e_t, y_t)= \alpha d((\bb x^c_s, \bb x^e_s), (\bb x^c_t, \bb x^e_t)) + \mathcal{L}(y_s, y_t).
    \label{eq:cost_ideal}
\end{align}
Since this cost function is symmetric, the transportation between the source and target domains is invertible.
However, 
$\bb x^e_s$ and $y_t$ are not observed in practice, making it necessary to estimate these values.
Although the label $y_t$ is substituted by its estimated value $\hat{y}_t = f(\bb x^c_t, \bb x^e_t)$,
the model to estimate the extra feature $\bb x^e_s$ is not considered in the proposed Algorithm~\ref{alg:alg1}.
Here, let us consider the estimation of $\bb x^e_s$.
A straightforward method of estimating $\bb x^e_s$ is to transfer the information of $\bb x^e$ in the target domain to the source domain by OT.
The cost function for this transportation is defined as 
\begin{equation}
    \mathcal{E}_\alpha^{ts}(\bb x^c_t, y_t; \bb x^c_s, y_s)=\alpha d(\bb x^c_t, \bb x^c_s) + \mathcal{L}(y_t, y_s).\label{eq:cost_func_ts}
\end{equation}
Owing to the lack of $\bb x^e_s$, the extra feature $\bb x^e$ is only considered with the estimated target label $f(\bb x^c_s, \bb x^e_t)$ that is used to substitute the unobserved target label $y_t$.
Here, an additional assumption is made so that the distribution of the extra features is estimated by OT using the above cost function, that is,
\begin{equation}
    \mathcal{P}_T(\bb x^e|\bb x^c, y) = \mathcal{P}_S(\bb x^e|\mathcal{T}(\bb x^c, y)),
    \label{eq:assumption_conditional}
\end{equation}
where $\mathcal{T}$ represents the OT of the common feature $\bb x^c$ and the label $y$ from the target distribution to the source distribution.
This assumption means that the conditional distribution of the extra feature $\bb x^e$ given $(\bb x^c, y)$ is identical before and after OT of the common feature $\bb x^c$ and the label $y$.
Under this assumption, the target extra features can be transferred to the source domain by OT.

After $\bb x^e_s$ is estimated using the above OT, it is possible to transfer the label information from the source domain to the target domain by OT using the cost function Eq.~\eqref{eq:cost_ideal}.
However, under the assumption Eq.~\eqref{eq:assumption_conditional}, there always exists a target sample that has the same extra feature $\bb x^e$ as a source sample at the destination of OT.
In other words, when we solve the OT problem for common features and labels, the transport cost of the extra features is always minimized to zero.
Therefore, eventually, the estimation of the source extra feature is not required, and solving the one-way OT problem Eq.~\eqref{eq:optimal_transport} is equivalent to solving the two-way OT problem.

As the above analysis shows, the two-way OT is reduced to one-way OT from the source domain to the target domain, and the extra features are ignored instead of estimating them in one-way OT.
This strategy is called \textit{cutoff} strategy in hybrid domain adaptation~\citep{8432087}.
Another simple strategy is to use a constant value, e.g., zero, for the unobserved source extra features. This strategy is called \textit{fill-up} strategy.
However, in the OT problem Eq.~\eqref{eq:optimal_transport} both strategies yield the same result when we use the $p$-th power of the $\ell_p$ distance as $d$.

When we use the fill-up strategy, the transport cost defined in Eq. \eqref{eq:cost_ideal} is used.
Let us consider the transport cost between the source samples $(\bb x^c_{si}, \hat{\bb x}^e_{si}, y_{si})\ (i=1,2,\ldots ,N_s)$ and the target samples $(\bb x^c_{tj}, \bb x^e_{tj}, \hat{y}_{tj})\ (j=1,2,\ldots ,N_t)$.
Here, assume that the source extra feature is estimated as $\hat{\bb x}^e_{si} = \gamma\bb 1$, where $\gamma\in\mathbb{R}$ is some constant value and $\bb 1$ is a vector of ones.
Namely, all of the source extra features are estimated as a constant value $c$.
When we use the $p$-th power of the $\ell_p$ distance as $d$, the transport cost is calculated as
\begin{align}
    \mathcal{E}_\alpha^*(\bb x^c_{si}, \hat{\bb x}^e_{si}, y_{si};\bb x^c_{tj}, \bb x^e_{tj}, \hat{y}_{tj})=\alpha(\|\bb x^c_{si}-\bb x^c_{tj}\|_p^p+\|c\bb 1-\bb x^e_{tj}\|_p^p) + \mathcal{L}(y_{si}, \hat{y}_{tj}).
    \label{eq:cost_constant}
\end{align}
In the above cost, the second term is common for all source samples for each target sample $j=1,2,\ldots ,N_t$.
Recall that the optimization problem of OT is
\begin{equation*}
 \minimize_{\substack{\hat{\pi}\in\hat{\Pi}(\mathcal{D}_S,\mathcal{D}_T^f)\\f\in\mathcal{F}}}\sum_{i=1}^{N_s}\sum_{j=1}^{N_t} \hat{\pi}_{ij}\mathcal{E}_\alpha^*(\bb x^c_{si},\hat{\bb x}^e_{si},y_{si};\bb x^c_{tj}, \bb x^e_{tj}, \hat{y}_{tj}),
\end{equation*}
which minimizes the total sum of the transport cost.
However, the second term of the cost in Eq.\eqref{eq:cost_constant} is equally added to the total sum of the cost for any transportation plan $\hat{\pi}$.
Therefore, for any $\gamma$, the same solution is obtained for the OT problem.
Furthermore, the cutoff strategy can be seen as a strategy that makes the second term of the cost in Eq.\eqref{eq:cost_constant} zero for any target sample.
In summary, our proposed method that considers the estimation of the source extra features is equivalent to both cutoff strategy and fill-in strategy.

\subsection{Learning Bound of Trained Model on Target Domain}
In this subsection, we show the learning bound of the model $f$ on the target domain.
The upper bound derived here is related to the upper bound derived in \citep{NIPS2017_0070d23b}.
Their upper bound focuses on the transportation between the source distribution and the estimated target distribution that is solved for JDOT.
However, their upper bound does not take into account the training of model $f$ involved in the practical algorithm.
On the other hand, the upper bound derived here focuses on training of the model $f$; hence, our upper bound becomes tighter with respect to the model.
More specifically, the derived upper bound consists of the Rademacher complexity of the model set and the Wasserstein distance between the estimated target distribution and the true target distribution.
We remark that although the Wasserstein distance between the estimated target distribution and the true target distribution is contained in the upper bound, the transportation between the source and estimated target distributions is not considered explicitly.
Namely, instead of considering the training of the model, our analysis does not consider how to estimate the target distribution explicitly.

To begin with, we consider the probabilistic transfer Lipschitzness introduced in \citep{NIPS2017_0070d23b}.
\begin{definition}[Probabilistic Transfer Lipschitzness]
Let $\mu_s$ and $\mu_t$ be the source and target distributions, respectively, and define $\phi(\lambda):\mathbb{R}\to[0, 1]$.
A labeling function $f:\Omega\to\mathbb{R}$ and a joint distribution $\pi(\mu_s, \mu_t)$ over the distributions $\mu_s$ and $\mu_t$ are $\phi$-Lipschitz transferable if for all $\lambda > 0$,
\begin{equation*}
        \Pr_{(\bb x_1, \bb x_2)\sim \pi(\mu_s, \mu_t)}[|f(\bb x_1)-f(\bb x_2)| > \lambda d(\bb x_1, \bb x_2)] \leq \phi(\lambda).
\end{equation*}
\end{definition}
The definition of probabilistic transfer Lipschitzness implies that if two instances $\bb x_1$ and $\bb x_2$ are sufficiently close,
the probability that these instances have different labels is bounded by $\phi(\lambda)$, where $\lambda$ is inversely proportional to the closeness of the instances.

To derive an upper bound, let $\mathcal{P}_{\hat{T}}$ be the distribution that estimates the true target distribution $\mathcal{P}_T$.
Then, we define the expected loss for a model $f\in\mathcal{F}$ with respect to each distribution $\mathcal{P}_T$,  $\mathcal{P}_{\hat{T}}$, as
\begin{align*}
    &\err_T(y, f)=\mathbb{E}_{(x^c, x^e, y)\sim \mathcal{P}_T} \mathcal{L}(y, f(x^c, x^e)),\\
    &\err_{\hat{T}}(y, f)=\mathbb{E}_{(x^c, x^e, y)\sim \mathcal{P}_{\hat{T}}} \mathcal{L}(y, f(x^c, x^e)).
\end{align*}
Also, let $\hat{\mathcal{P}}_{\hat{T}}$ be the empirical distribution that consists of samples $\{(\bb x^c_i, \bb x^e_i, y_i)\}_{i=1}^{N_t}$ that follow $\mathcal{P}_{\hat{T}}$.
Then, the empirical loss for model $f$ with respect to $\hat{\mathcal{P}}_{\hat{T}}$ is defined as
\begin{align*}
    \widehat{\err}_{\hat{T}}(y, f)=\frac{1}{m}\sum_{i=1}^{m} \mathcal{L}(y_i, f(\bb x^c_i, \bb x^e_i)).
\end{align*}
Let $f_0$ and $\hat{f}_s$ be the models that minimize the expected loss $\err_T(y, f)$ and the empirical loss $\widehat{\err}_{\hat{T}}(y, f)$:
\begin{align*}
    &f_0 = \inf_{f\in\mathcal{F}} \err_T(y, f),\\
    &\hat{f}_s = \min_{f\in\mathcal{F}} \widehat{\err}_{\hat{T}}(y, f).
\end{align*}

Now, we are ready to present our main result. 
Assume that the following condition holds.
\begin{itemize}
\item The space of the target features, $\Omega_t$, is endowed with a positive definite kernel $K$, and let $\mathcal{H}$ be its associated reproducing kernel Hilbert space.
\item The kernel $K$ is bounded as $\sup_{\bb x\in\Omega_t} K(\bb x, \bb x)\leq \Lambda^2$.
\item The model set $\mathcal{F}$ is a ball of radius $a$ in $\mathcal{H}$, namely, $\mathcal{F}=\{f\in\mathcal{H}\mid \|f\|_\mathcal{H}\leq a\}$.
\item The loss function $\mathcal{L}$
\begin{itemize}
    \item is symmetric, $\mathcal{L}(y_1, y_2)=\mathcal{L}(y_2, y_1)$,
    \item satisfies the triangle inequality, $\mathcal{L}(y_1,y_2)+\mathcal{L}(y_2, y_3)\geq \mathcal{L}(y_1, y_3)$,
    \item is Lipschitz continuous with constant $k$, $|\mathcal{L}(y_1, y_2)-\mathcal{L}(y_1, y_3)|\leq k|y_2-y_3|$, and
    \item is bounded as $L_0=\sup_{y\in\mathcal{C}}\mathcal{L}(0, y)<\infty$.
\end{itemize}
\item The optimal model $f_0\in\mathcal{F}_0$ is upper bounded as, for all $\bb x_1^c, \bb x_1^e, \bb x_2^c, \bb x_2^e$, $|f_0(\bb x^c_1,\bb x^e_1)-f_0(\bb x^c_2,\bb x^e_2)|\leq M$.
\item The optimal model $f_0$ and the OT plan $\pi^*$ from $\mathcal{P}_{\hat{T}}$ to $\mathcal{P}_T$ satisfy the $\phi$-probabilistic transfer Lipschitzness.
\end{itemize}
We remark that $\mathcal{F}\neq\mathcal{F}_0$ in general.
Then, our main result is summarized as follows.
\begin{theorem}
\label{thm:main_theorem}
Under the above assumptions, let $\hat{f}_s\in\mathcal{F}$ be the trained model that minimizes the empirical loss $\widehat{\err}_{\hat{T}}$ of the distribution $\mathcal{P}_{\hat{T}}$ that estimates the true target distribution.
Then, for all $\lambda > 0$ with $\alpha=k\lambda$, for any $\delta\in(0, 1)$ with probability at least $1-\delta$,
\begin{align*}
    \err_T(y,\hat{f}_s)\leq&\widehat{\err}_{\hat{T}}(\hat{f}_s, y) +\frac{2ka\Lambda}{\sqrt{N_t}}+ (L_0 + ka\Lambda)\sqrt{\frac{\ln(1/\delta)}{N_t}}+W(\mathcal{P}_{\hat{T}},\mathcal{P}_T)\\
    &+2\err_T(y, f_0)+ kM\phi(\lambda),
\end{align*}
where $W(\mathcal{P}_{\hat{T}},\mathcal{P}_T)$ is the Wasserstein distance between the estimated target distribution and the true target distribution and is defined as
\begin{align*}
&W(\mathcal{P}_{\hat{T}},\mathcal{P}_T)\\&= \inf_{\pi\in\Pi(\mathcal{P}_{\hat{T}}, \mathcal{P}_T)}\int_{(\Omega_t\times \mathcal{C})^2}\mathcal{E}^*_\alpha(\bb x^c_1, \bb x^e_1, y_1; \bb x^c_2, \bb x^e_2, y_2)d\Pi(\bb x^c_1, \bb x^e_1, y_1;\bb x^c_2, \bb x^e_2, y_2).
\end{align*}
\end{theorem}
The detailed proof of Theorem~\ref{thm:main_theorem} is presented in the appendix.
When we assume $\mathcal{P}_{\hat{T}}=\mathcal{P}_T^f$, the upper bound corresponds to Algorithm~\ref{alg:alg1}.
Examples of the loss functions that satisfy the above assumptions are the 0-1 loss function for a classification problem and the $\ell_p$ distance on a finite set for a regression problem.
The above upper bound is divided into three parts.

The first part, $\widehat{\err}_{\hat{T}}(\hat{f}_s, y)+\frac{2ka\Lambda}{\sqrt{N_t}}+(L_0+ka\Lambda)\sqrt{\ln(1/\delta)/N_t}$, is an upper bound based on the Rademacher complexity and estimates $\err_{\hat{T}}(\hat{f}_s, y)$ from a finite number of samples that follow $\mathcal{P}_{\hat{T}}$.
In addition, $\hat{f}_s$ is obtained by minimizing $\widehat{\err}_{\hat{T}}$; hence, these terms are minimized in terms of the model $f\in\mathcal{F}$.

The second part, $W(\mathcal{P}_{\hat{T}},\mathcal{P}_T)$, is the discrepancy between the estimated and true target distributions.
This distance becomes small if the estimated target distribution is close to the true target distribution irrespective of the distance between the source and target distributions. 
Namely, this term focuses on the transferability of the source label information to the target domain. 
In general, theoretical analyses of domain adaptation such as~\cite{Ben-David2010,redko:hal-01613564,NIPS2017_0070d23b} evaluates the discrepancy between the source and target distributions.
However, the source and target distribution have different dimensionality in our problem, and there are no clear definition of the discrepancy between the distributions of different dimensionalities.
In addition, the simple idea such as using the discrepancy between the embedded distributions does not properly evaluate the transferability of source label information.
Instead, here we evaluate the Wasserstein distance of the true and estimated target distributions, which have the same dimensionality, and focus on the bound of the training of the model using estimated target distribution.
Further detailed analysis to evaluate the source and target discrepancy for our problem is left for future work.

The last part, $2\err_T(y, f_0) + kM\phi(\lambda)$, is determined by the predictability of the target distribution and the probabilistic transfer Lipschitzness of model $f_0$; hence, these terms are considered constants that depend on the problem.
In conclusion, the upper bound becomes tight when the estimated and true target distributions are close, and the model can accurately predict the transferred label.

\section{Numerical Experiments}
\label{sec:experiments}
In this section, we present experimental results of domain adaptation problems for the observation of extra features using both synthetic and real data.
Our code used in the following experiments is based on Python Optimal Transport (POT)~\citep{flamary2021pot}, and is publicly available at \url{https://github.com/t-aritake/DAEVS}.

\subsection{Experiments with Synthetic Data}
\begin{figure}[b!]
    \begin{minipage}[b]{0.48\linewidth}
        \centering
        \includegraphics[width=\textwidth]{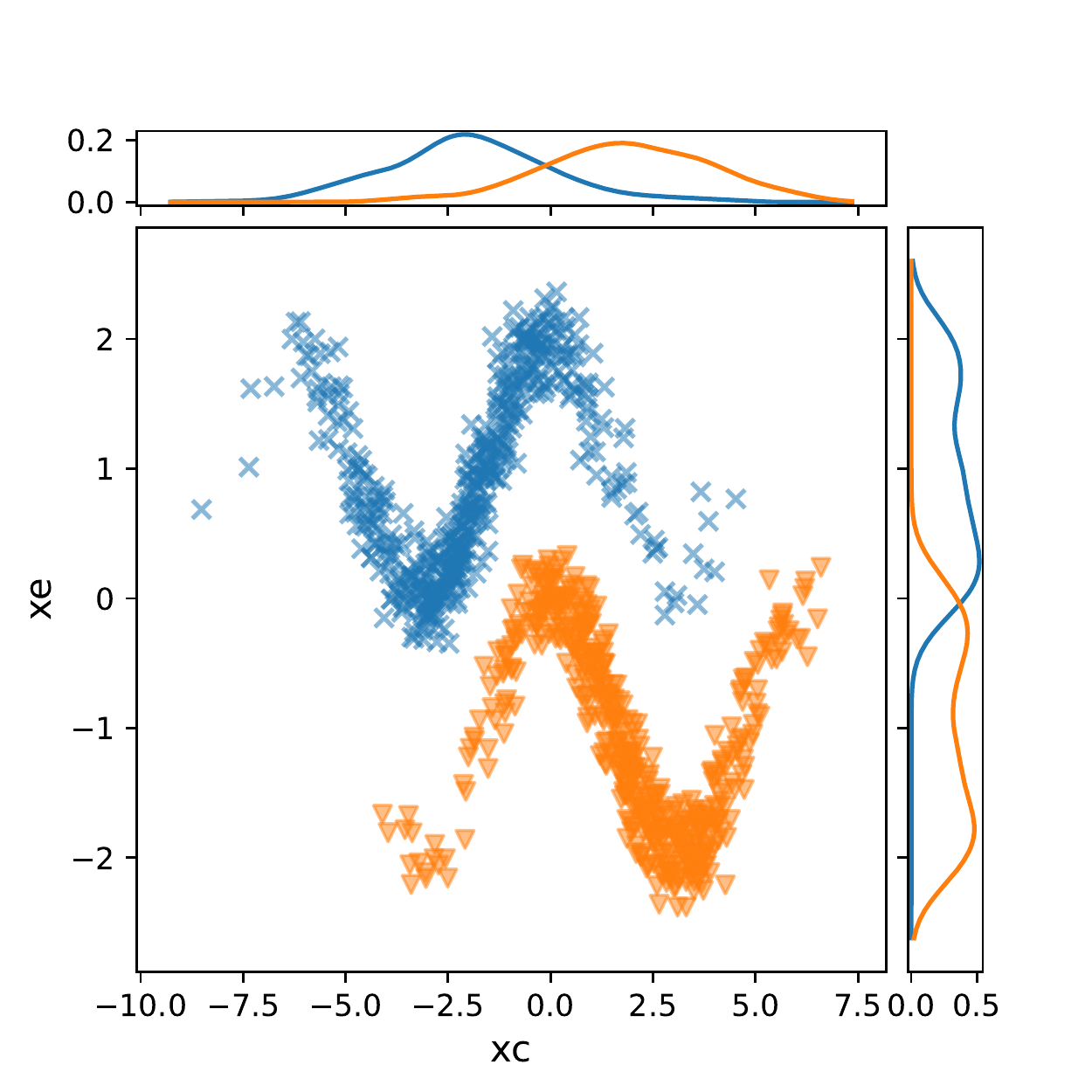}
        \subcaption{Source dataset}\label{fig:dataset_example_source}
    \end{minipage}
      \begin{minipage}[b]{0.48\linewidth}
        \centering
        \includegraphics[width=\textwidth]{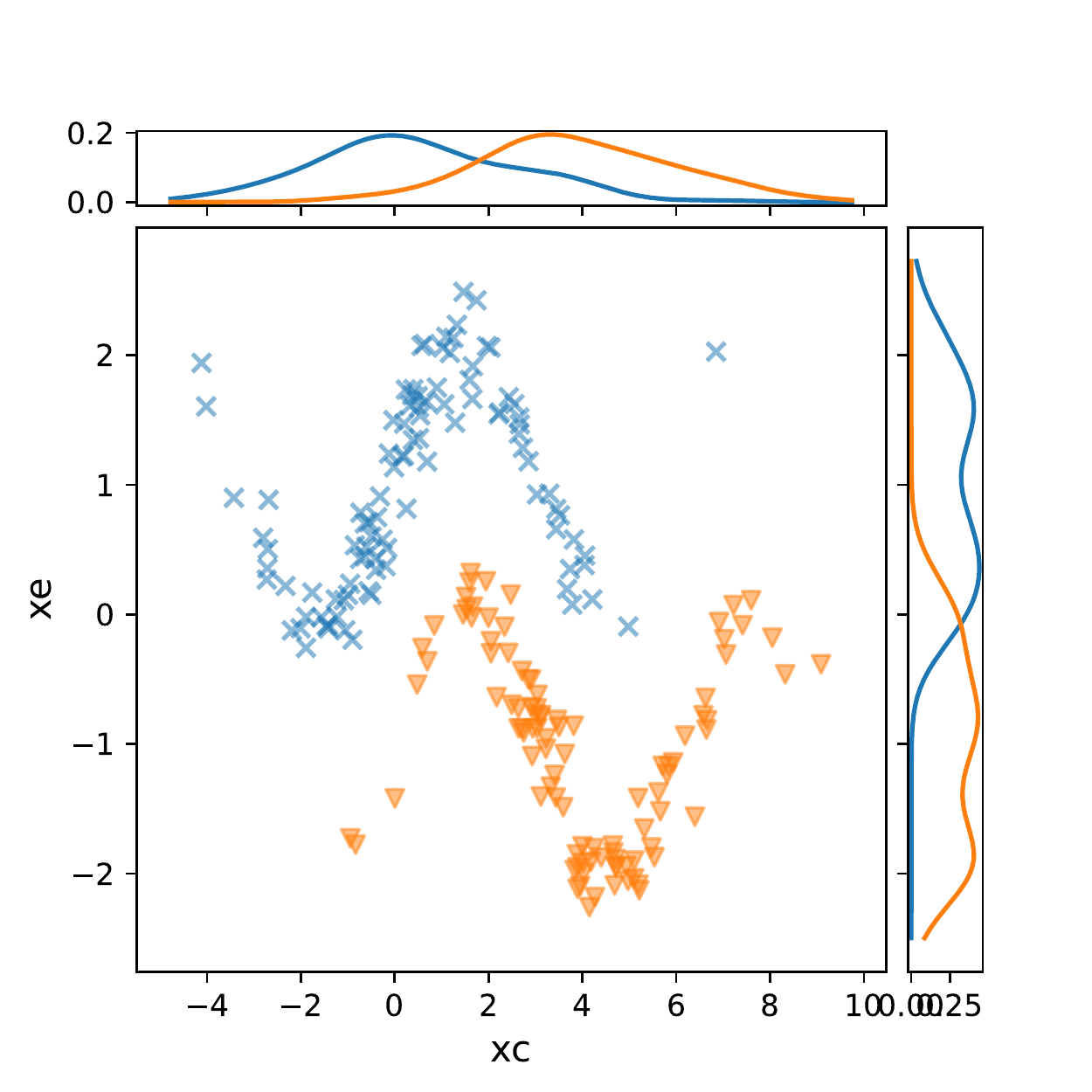}
        \subcaption{Target dataset}\label{fig:dataset_example_target}
      \end{minipage}
      \caption{Example of the true dataset used in the experiment. Triangles show positive samples and crosses show negative samples. The center of each class is different in the source and target domains, and the shape of the distribution is slightly different. The $x_e$ of the source samples and the labels of the target samples are not accessible in the experiment.}\label{fig:dataset_example}
\end{figure}

In this subsection, we show experimental results obtained with synthetic data.
In this experiment, unsupervised domain adaptation for a binary classification problem is considered.
We assume the dataset shown in Fig.~\ref{fig:dataset_example} as the true source and target datasets.
Here, both the common and extra features are one-dimensional for the purpose of visualization, and
the extra feature in the source domain and the labels in the target domain are not used for the classification.
We set the number of samples to be $N_s=1,000$ and $N_t=100$ for the source and target domains, respectively.
The dataset of each domain contains the same number of positive and negative samples.

We compared our proposed method with JDOT~\citep{NIPS2017_0070d23b}, which ignores the extra feature $x^e$.
Also, JDOT in the ideal situation, where the extra feature is observed in both source and target domains, is used as a benchmark for the optimal performance.
We assume that the class set $\mathcal{F}$ is the set of support vector machines (SVMs) with a Gaussian kernel.
We use the training loss of SVMs as $\mathcal{L}$, which is a surrogate loss of 0-1 loss, and set the balancing parameter $\alpha$ of Eq.~\eqref{eq:transportation_cost} to $\alpha=1$, which was experimentally determined.
The effect of the choice of the model set and balancing parameter $\alpha$ is further discussed in the appendix.

\begin{figure}[b!]
    \centering
    \includegraphics[width=.5\textwidth]{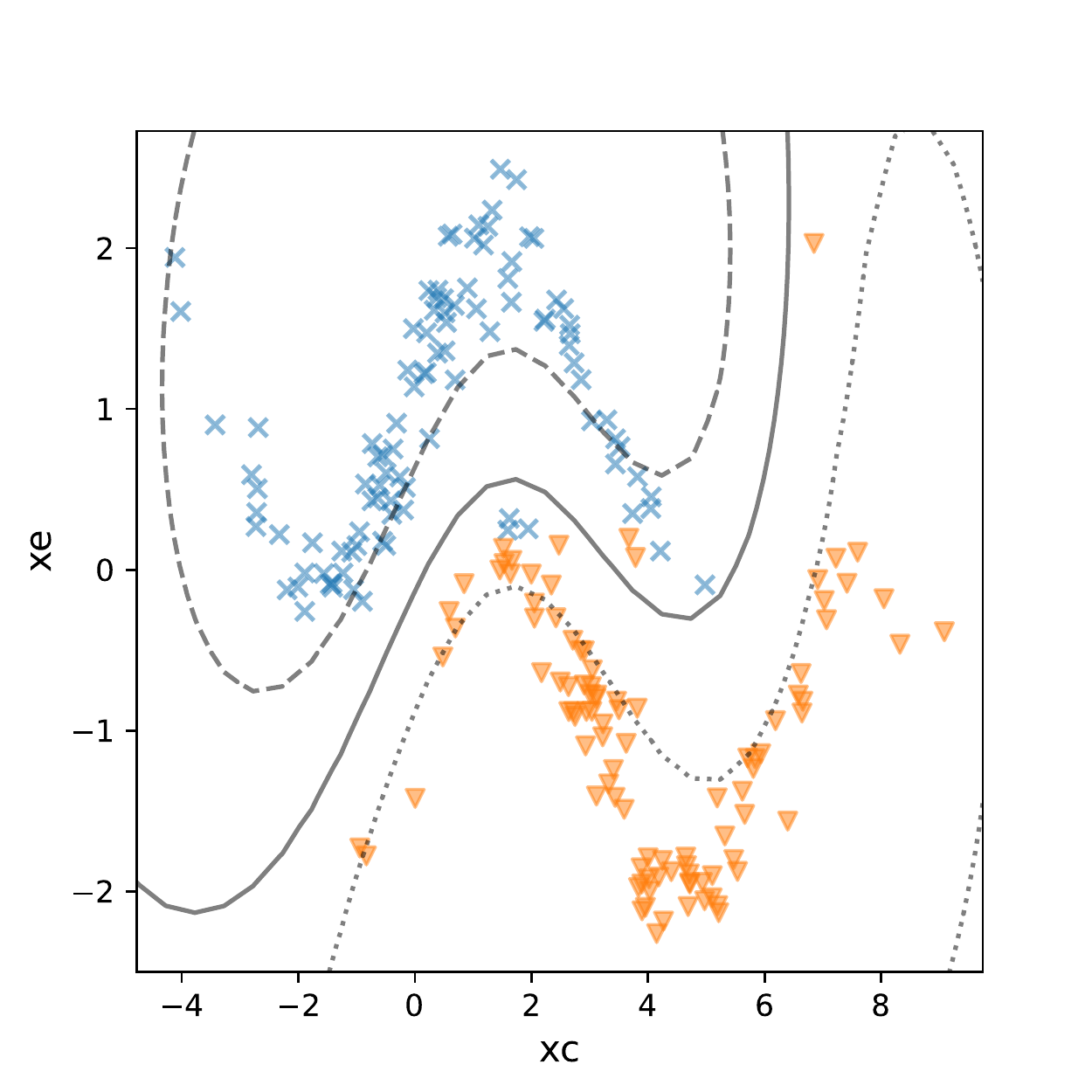}
    \caption{Transferred labels in the target domain and the obtained decision boundary of the learned model.}
    \label{fig:test_boundary}
\end{figure}

The classification accuracy for the proposed method, JDOT ignoring the extra feature, and optimal benchmark is shown in Table~\ref{tab:synthetic_results}.
We generated 10 different random datasets, where each dataset is similar to the dataset in Fig.~\ref{fig:dataset_example}.
Then, we calculated the average prediction accuracy and its variance of the transferred label (transfer) and the estimated target label using the trained model $f$ (model).
We evaluated these values because it is possible to build an accurate model from partly incorrectly transferred labels, or conversely, there is possibility to build an inaccurate model from the correctly transferred labels.
Also, Fig.~\ref{fig:test_boundary} shows the decision boundary of a trained model in the target domain obtained by the proposed method.
\begin{table*}[b!]
    \small
    \centering
    \caption{Accuracy with synthetic data}
    \begin{tabular}{P{8ex}P{2.2cm}P{2.2cm}P{2.2cm}P{2.2cm}|P{2.2cm}}
        \hline
         & Proposed & JDOT\hspace{6ex}no extra&CCA & DSFT & JDOT\hspace{6ex}ideal\\
         \hline
        transfer & \textbf{94.0 } ($7.78\times 10^{-2}$) & 79.7 ($1.68\times 10^{-2}$) & ---& ---& 95.3 ($2.53\times 10^{-2}$)\\
        model& \textbf{95.5} ($6.80\times 10^{-2}$) & 80.3 ($2.51\times 10^{-2}$) & 52.6 ($8.81\times 10^{-2}$) &70.8 ($3.09\times 10^{-2}$)& 96.4 ($2.29\times 10^{-2}$)\\
        \hline
    \end{tabular}
    \label{tab:synthetic_results}
\end{table*}

From Table~\ref{tab:synthetic_results}, we can see that our proposed method consistently outperforms JDOT without an extra feature, and the model accuracy is higher than the transfer accuracy in both methods.
The reason that our proposed method outperforms JDOT is that the marginal distributions of $x^c$ and $x^e$ of the positive class and those of the negative class are highly overlapped, as can be seen in Fig.~\ref{fig:dataset_example}.
Therefore, it is difficult to build a model that accurately predicts the label only from the common features.
On the other hand, by considering both the common feature and the extra feature for the OT, our proposed method accurately estimates the true target distribution.
Furthermore, as shown in Fig.~\ref{fig:test_boundary}, even when some of the labels are not correctly transferred, the trained model is able to estimate the true target labels accurately.
Therefore, the test accuracy of the trained model outperforms the accuracy of the OT.
Note, however, that this result depends on the complexity of the model, and we conjecture that the non-linearity of the ground-truth decision boundary affects the possibility of domain adaptation for our problem.
The effects of the choice of the model set $\mathcal{F}$ or the parameter $\alpha$ are discussed in the appendix.
Here, we provide qualitative analysis for the successful domain adaptation for our problem. 

\begin{figure}[p]
    \centering
    \begin{minipage}[t]{0.45\linewidth}
        \centering
        \includegraphics[width=\textwidth]{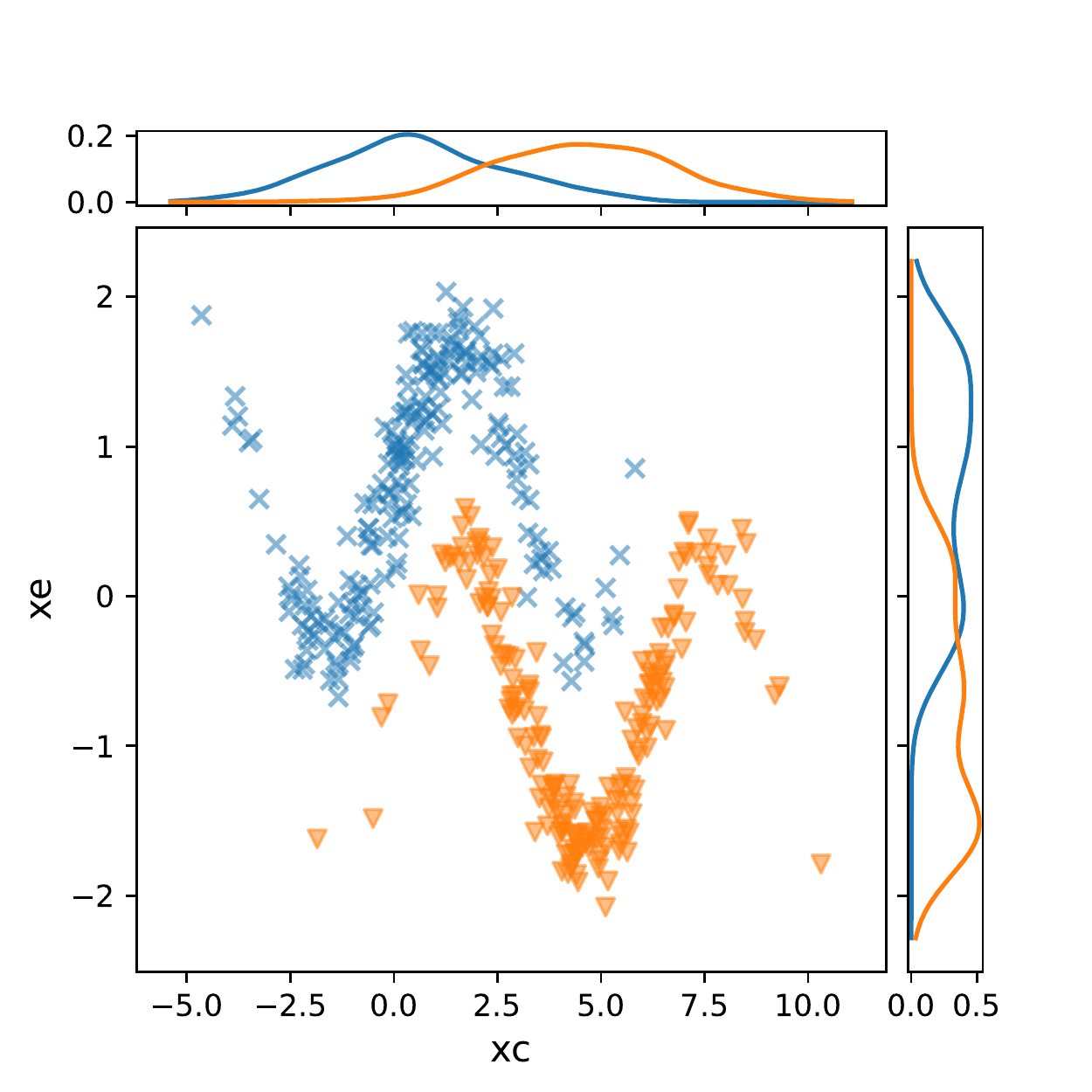}
        \subcaption{Overlapped target distribution}
    \end{minipage}
    \hspace{2ex}
      \begin{minipage}[t]{0.45\linewidth}
        \centering
        \includegraphics[width=\textwidth]{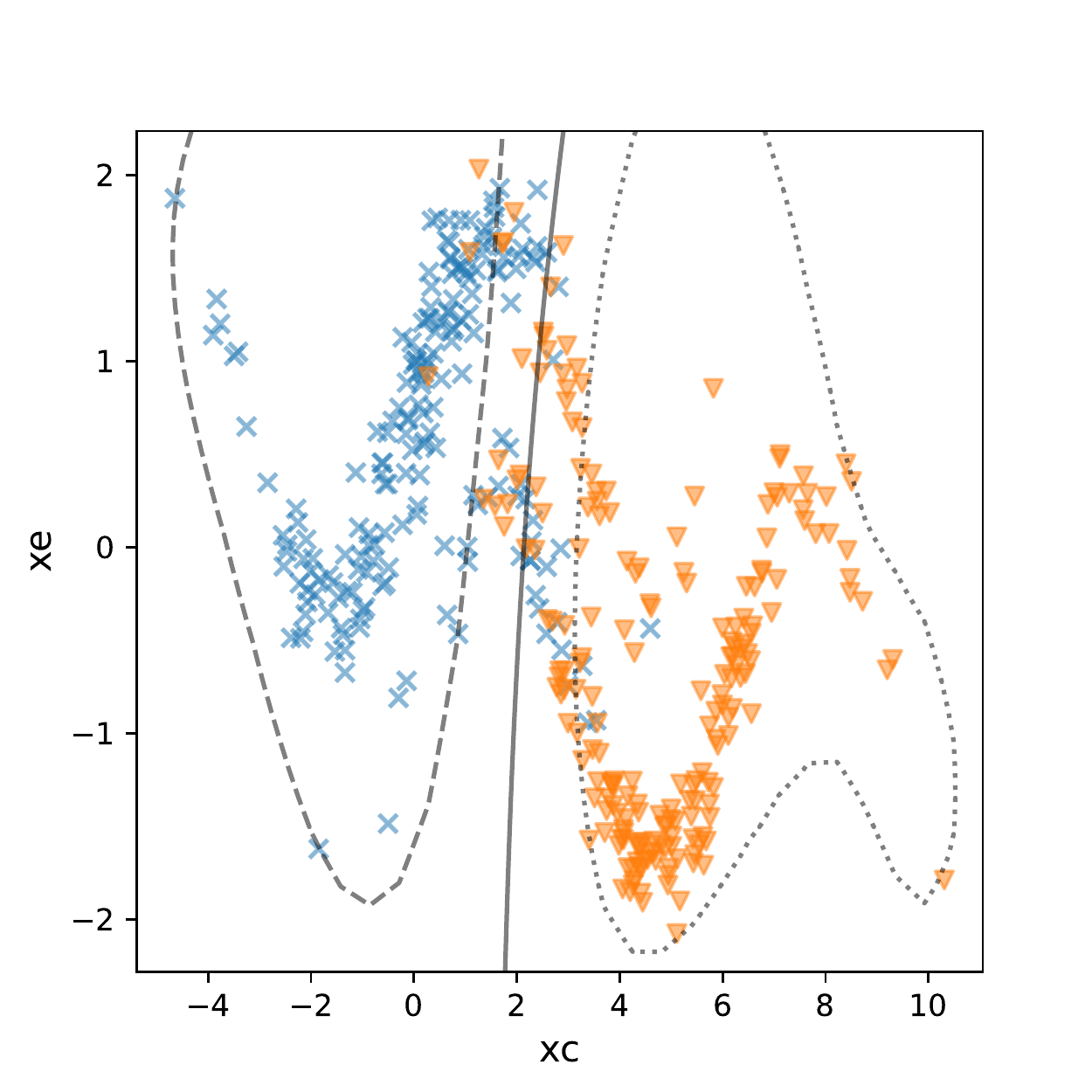}
        \subcaption{Estimated target distribution by OT and initial model learned from the estimated distribution}
      \end{minipage}
    \caption{(a) An example of overlapped target distribution and (b) the initial label distribution estimated by OT using only distance between samples for the transportation costs.} 
    \label{fig:overlapped_distributions}
\end{figure}

More specifically, we conjecture that the linearity of the ground-truth decision boundary with respect to extra features $\bb x^e$ is important for the successful domain adaptation when the distributions of different classes overlap in the space of the source common features.
Our proposed method transfer source label information based on the distance between common features and the prediction error.
Therefore, when the prediction model is sufficiently close to the true decision boundary, it is possible to estimate the target labels accurately.
However, at the early steps of the proposed algorithm, the prediction model is not in general close to the true decision boundary, because only the distance between the common features is used as the cost for OT for initial label assignment.
Figure~\ref{fig:overlapped_distributions} (a) shows an example of a distribution where the distributions of different classes are overlapped in the common feature space.
As we can see in Figure~\ref{fig:overlapped_distributions} (b), the labels of overlapped region which are assigned by initial OT is determined according the ratio of the class label for a given common features.
The initial model obtained from such labels is not close to the true decision boundary as shown in Figure~\ref{fig:overlapped_distributions} (b). 
Therefore, the success of proposed method depends on whether the prediction model that are close to the true labeling function is obtained after several iterations of the proposed algorithm.
\begin{figure}[p]
    \centering
    \begin{minipage}[t]{0.45\linewidth}
        \centering
        \includegraphics[width=\textwidth]{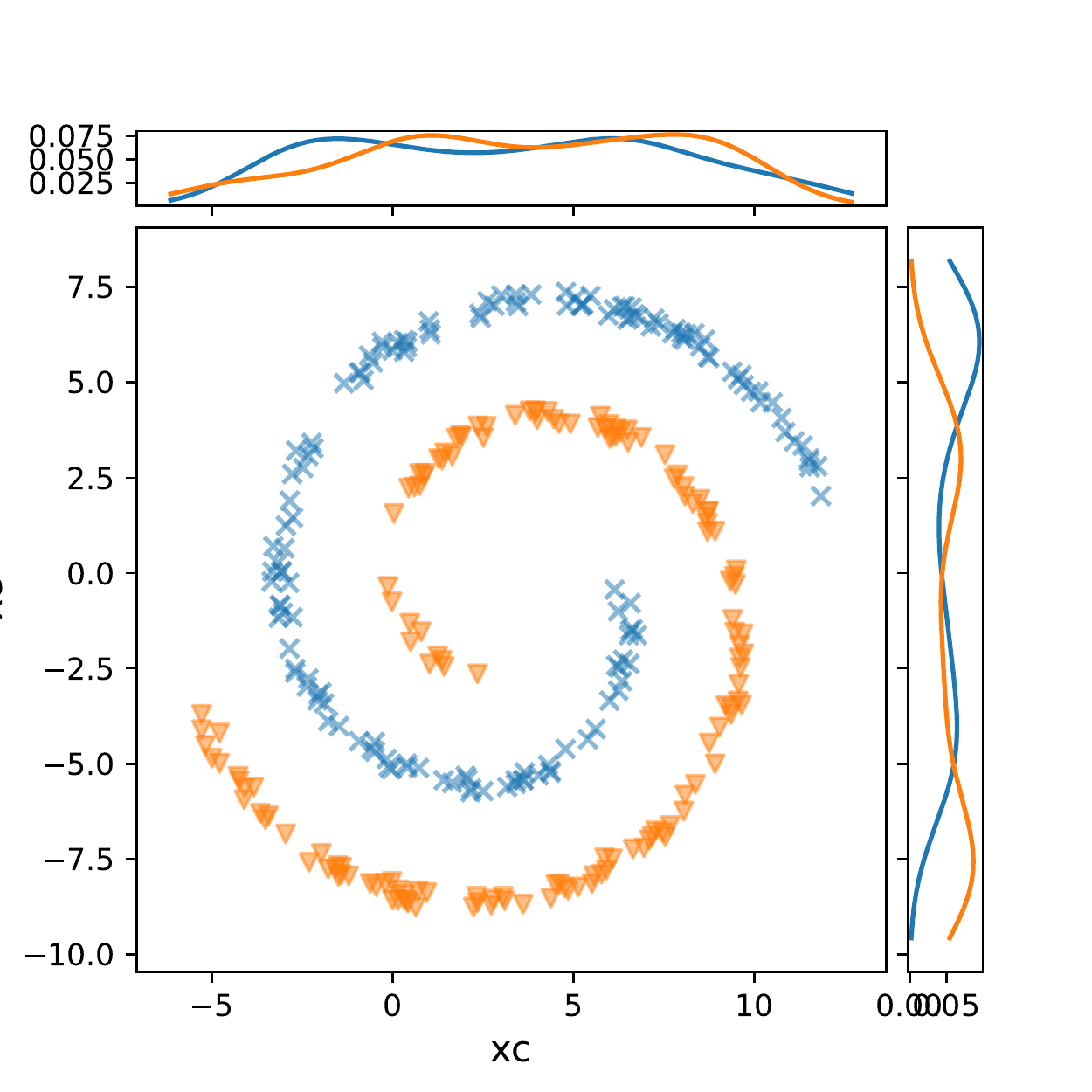}
        \subcaption{Target distribution}
    \end{minipage}
    \hspace{2ex}
      \begin{minipage}[t]{0.45\linewidth}
        \centering
        \includegraphics[width=\textwidth]{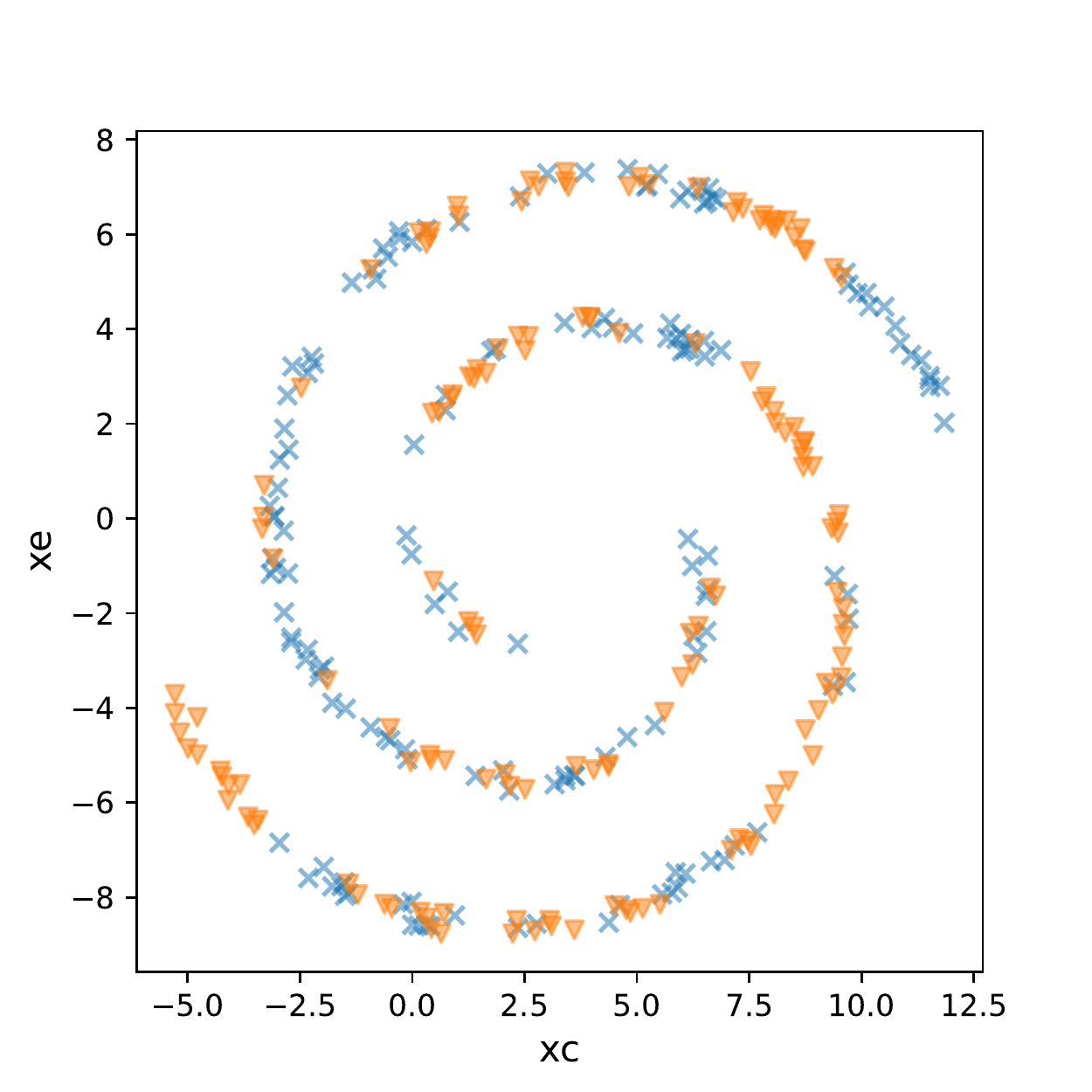}
        \subcaption{Estimated initial target distribution by OT}
      \end{minipage}
    \caption{(a) An example of overlapped target distribution whose labeling function is nonlinear function of $x^e$ and (b) the initial label distribution estimated by OT.}
    \label{fig:overlapped_distributions_spirals}
\end{figure}

When the true labeling function $F(\bb x^c, \bb x^e)=0$ is linear in terms of the extra feature $\bb x^e$, the prediction model close to $F$ can be obtained after several iterations.
For example, the decision function of Figure~\ref{fig:overlapped_distributions} is $F(x^c, x^e)=\cos(x^c)-x^e=0$, which is linear in terms of $x^e$.
In this case, although the tentative models of early steps is not close to the true labeling function,
the labels given a common feature close to the tentative decision boundary can be modified so that the samples in the target domain is divided into two clusters by considering the prediction error of the tentative model as the cost for OT.
However, when $F$ is nonlinear function of $\bb x^e$, target samples given a common variable $\bb x^c$ cannot be linearly separable.
as shown in Figure~\ref{fig:overlapped_distributions_spirals}.
In this case, the labels of the target samples cannot be correctly modified using the tentative models of early steps.
method would probably be failed, because the complex structure of the target distribution cannot be captured by the prediction model.

\subsection{Experiments With Real Data}
In this subsection, we show experimental results obtained with real data.
We used the gas sensor array drift dataset used in~\citep{VERGARA2012320}.
The original data are 16-channel time series obtained by measuring one of six gases at different concentration levels using an array of 16 gas sensors.
The dataset consists of 10 batches, where the samples in each batch are obtained for a different month and are affected by different levels of sensor drift; hence, each batch can be used as a dataset of different domains.
We used the first four batches and considered a domain adaptation problem between these batches.
Also, we consider the binary classification problem to classify only two types of gases, ethanol and ethylene, out of the six types of gases.
We used the six transient features extracted from each sensor for classification.
We selected eight out of 16 sensors, and transient features extracted from the selected sensors are used as common features, while the features extracted from the rest of the sensors are used as extra features.
Here, the sensors used to extract common features are selected so that the extra features make the classification more accurate.
Although, in practice, it is possible that the extra features do not contribute to the accuracy of the classification, here, we considered the reasonable scenario that informative features for the classification are observed as extra features in the target domain.

Table~\ref{tab:results_real} shows the prediction accuracy in the target domain for each domain adaptation problem.
The row of domains $A\to B$ shows the experimental results where batch~$A$ and batch~$B$ are used as the source and target domains, respectively.
The Baseline column shows the prediction accuracy on the test data without domain adaptation.
Namely, the baseline model is learned using only common features given in the source domain.
Similarly, the model accuracies of JDOT and the proposed method are shown in the table.
As we can see from the table, some domains do not require domain adaptation, and the baseline model outperforms JDOT and the proposed method.
However, for other domains, the prediction accuracy is largely improved by considering domain adaptation by OT.
In addition, the estimation accuracy of the proposed method exceeds that of JDOT in most domains using the informative extra features.
Other results with real data are presented in the appendix.
\begin{table}[t!]
    \small
    \centering
    \caption{Accuracy for real data}
    \begin{tabular}{P{8ex}cP{10ex}P{8ex}P{6ex}c|P{8ex}}
        \hline
         domains & Baseline & JDOT\ \ no extra& CCA & DSFT & Proposed & JDOT ideal\\
         \hline
         1$\to$ 2 & \textbf{83.33} & 77.71 & 66.87 & 42.97 & 78.31 & 83.73\\
         1$\to$ 3 & 52.28 & 93.45 & 43.27 & 57.78 & \textbf{96.02} & 94.15\\
         1$\to$ 4 & 64.49 & 60.75 & 58.88 & \textbf{94.39} & 87.85 & 85.98\\
         2$\to$ 1 & 52.13 & 79.26 & 56.91 & 62.77 & \textbf{84.04} & 85.64\\
         2$\to$ 3 & 56.84 & 89.36 & 25.03 & 84.56 & \textbf{89.47} & 90.99\\
         2$\to$ 4 & 63.55 & 69.16 & 51.40 & 41.12 & \textbf{71.96} & 74.77\\
         3$\to$ 1 & 51.06 & 92.02 & 92.55 & 66.49 & \textbf{94.15} & 95.74\\
         3$\to$ 2 & 68.67 & 81.92 & 67.67 & 86.94 & \textbf{88.76} & 88.55\\
         3$\to$ 4 & 94.39 & 77.57 & \textbf{96.26} & 40.19 & 81.31 & 80.37\\
         4$\to$ 1 & 50.00 & 52.66 & 48.93 & 51.60 & \textbf{53.72} & 80.85\\
         4$\to$ 2 & 42.97 & 71.08 & 32.93 & 32.93 & \textbf{74.30} & 73.89\\
         4$\to$ 3 &\textbf{92.98} & 82.81 & 57.31 & 42.69 & 82.57 & 82.57\\
         \hline
    \end{tabular}
    \label{tab:results_real}
\end{table}
\section{Conclusion}
\label{sec:conclusion}
In this paper, we considered the domain adaptation problem in which common features are  observed in both the source and target domains, and extra features are observed only in the target domain.
We proposed an unsupervised domain adaptation method for this extended feature space based on OT.
We showed that the OT of the proposed method is equivalent to the two-way OT between the domains under the assumption that the conditional distribution of extra features given common features and labels is identical before and after OT.
Also, we derived a learning bound of the model in the target domain on the basis of the Rademacher complexity and the Wasserstein distance between the estimated and true target distributions.
The experimental results demonstrate the ability to estimate a distribution close to the true target distribution by the proposed method with the accurate estimation of the target labels using the trained model.

The accurate estimation of the true target distribution is not always possible, and the conditions for the success of the estimation by the proposed method are not yet fully understood. The analysis of such conditions is important future work.
Furthermore, the case where some of the features in the source domain become unobservable (because of, e.g., mechanical breakdown of sensors) should also be discussed as a future extension of the proposed method.

\section*{Acknowledgement}
Part of this work is supported by JST CREST JPMJCR1761, JPMJCR2015, Mirai Project JPMJMI21G2, and JSPS JP20K06922.

\appendix
\counterwithin{figure}{section}
\counterwithin{equation}{section}
\counterwithin{table}{section}
\renewcommand{\thetheoremA}{A\arabic{theoremA}}
\renewcommand{\theequation}{A\arabic{equation}}
\renewcommand{\thetable}{A\arabic{table}}
\renewcommand{\thefigure}{A\arabic{figure}}
\section{PROOFS}
\subsection{Proof of Theorem 1}
In this section, we present the detailed proof of Theorem 1.
We recall the definitions of the distributions and the losses used in the following proof.
Let $\mathcal{P}_T$ be the true target distribution and $\mathcal{P}_{\hat{T}}$ be the distribution that estimates $\mathcal{P}_T$.
Then, the expected losses for a model $f\in\mathcal{F}$ with respect to $\mathcal{P}_T$ and $\mathcal{P}_{\hat{T}}$ are defined as
\begin{align*}
    &\err_T(y, f)=\mathbb{E}_{(x^c, x^e, y)\sim \mathcal{P}_T} \mathcal{L}(y, f(x^c, x^e)),\\
    &\err_{\hat{T}}(y, f)=\mathbb{E}_{(x^c, x^e, y)\sim \mathcal{P}_{\hat{T}}} \mathcal{L}(y, f(x^c, x^e)),
\end{align*}
respectively, where $\mathcal{L}$ is a classification loss function.
Then, the empirical loss for a model $f$ with respect to $\hat{\mathcal{P}}_{\hat{T}}$ is defined as
\begin{align*}
    \widehat{\err}_{\hat{T}}(y, f)=\frac{1}{m}\sum_{i=1}^{m} \mathcal{L}(y_i, f(\bb x^c_i, \bb x^e_i)).
\end{align*}
Then, let $f_0$ and $\hat{f}_s$ be the models that minimize the expected loss $\err_T(y, f)$ and the empirical loss $\widehat{\err}_{\hat{T}}(y, f)$:
\begin{align*}
    &f_0 = \inf_{f\in\mathcal{F}} \err_T(y, f),\\
    &\hat{f}_s = \min_{f\in\mathcal{F}} \widehat{\err}_{\hat{T}}(y, f).
\end{align*}
Here, the model $f_0$ is the ideal model that minimizes the true target loss and $\hat{f}_s$ is the model that is obtained from the samples that follow estimated target distribution.

First, we consider the difference of the expected loss $\err_T(y, \hat{f}_s)-\err_T(y, f_0)$.
Since the model $f_0$ minimizes the expected loss in the target domain $\err_T(y, f)$, this difference evaluates the deviation of the model $\hat{f}_s$ from the ideal model $f_0$.
For this difference, the following inequality holds.
\begin{align}
    \err_T(y, \hat{f}_s)-\err_T(y, f_0) &\leq \err_T(\hat{f}_s, f_0)\nonumber\\
    &=\err_{\hat{T}} (\hat{f}_s, f_0)\nonumber\\
    &\leq \err_{\hat{T}}(y, \hat{f}_s) + \err_{\hat{T}}(y, f_0)\label{eq:triangular1}.
\end{align}
The first and the last line follows from the definition of the expected loss and the assumption that the loss function $\mathcal{L}$ is symmetric and satisfies the triangular inequality.
The second line also follows from the definition of $\err_T$,
\begin{equation*}
    \err_T(\hat{f}_s, f_0)=\mathbb{E}_{(\bb x^c, \bb x^e, y)\sim \mathcal{P}_T} \mathcal{L}(\hat{f}_s(\bb x^c, \bb x^e), f(\bb x^c, \bb x^e)).
\end{equation*}
It is obvious from the above equation that the distribution of $y$ does not affect the $\err_T(\hat{f}_s, f_0)$; hence, any distribution that has the same marginal distribution of $(\bb x^c, \bb x^e)$ as $\mathcal{P}_T$ can be used in place of $\mathcal{P}_T$ to calculate $\err_T(\hat{f}_s, f_0)$.
Then, we consider the upper bound of the two terms of the right-hand side of Eq.~\eqref{eq:triangular1}.

First, we consider the upper bound of the first term of the right-hand side of Eq.~\eqref{eq:triangular1}.
Here, we use the following uniform law of large numbers~\cite{vapnik1998statistical}.
\begin{theoremA}[Uniform law of large numbers]
\label{thm:ULLN}
Suppose $\mathcal{G}\subset\{g:\mathcal{Z}\to[b_l, b_h]\}$.
Let $Z_1,\ldots ,Z_n$ be independent and identically distributed random features of the distribution $D$, where the random feature $Z$ also follows $D$.
Then, for all $\delta\in(0, 1)$, with probability at least $1-\delta$,
\begin{equation}
\sup_{g\in\mathcal{G}} \left(\mathbb{E}[g(Z)]-\frac{1}{n}\sum_{i=1}^n g(Z_i)\right)\leq
2\mathfrak{R}_n(\mathcal{G}) + (b_h-b_l)\sqrt{\frac{\ln (1/\delta)}{2n}},
\label{eq:uniform_law_of_large_numbers}
\end{equation}
where $\mathfrak{R}_n(\mathcal{G})$ is the empirical Rademacher complexity calculated from $n$ samples.
\end{theoremA}
We can apply this theorem with $\mathcal{G}=\mathcal{L}\circ \mathcal{F}\equiv\{\mathcal{L}\circ f \mid f\in\mathcal{F}\}$, $Z=(\bb X^c, \bb X^e, Y)$, and $Z_i = (\bb X^c_i, \bb X^e_i, Y_i)\ (i=1,2,\ldots ,N_t)$ and $D=\mathcal{P}_{\hat{T}}$.
Here, we define $g(Z) = \mathcal{L}\circ f(Z)\equiv\mathcal{L}(Y, f(\bb X^c, \bb X^e))$ for $f \in \mathcal{F}$.
In addition, recall the assumption that $\mathcal{F}=\{f\in\mathcal{H}\mid\|f\|_\mathcal{H}\leq a\}$ where $\mathcal{H}$ is a reproducing kernel Hilbert space whose kernel $K$ is a bounded kernel with $\sup_{x}K(\bb x, \bb x)=\Lambda^2 <\infty$. Then,
\begin{align}
    &\mathbb{E}_{(\bb X^c, \bb X^e, Y)\sim\mathcal{P}_{\hat{T}}}[g(Z)] = \mathbb{E}_{(\bb X^c, \bb X^e, Y)\sim\mathcal{P}_{\hat{T}}}[\mathcal{L}(Y, f(\bb X^c, \bb X^e))] = \err_{\hat{T}}(y, f),\label{eq:err_hatt}\\
    &\frac{1}{N_t}\sum_{i=1}^{N_t} g(Z_i) = \frac{1}{N_t}\mathcal{L}(Y_i, f(\bb X^c_i, \bb X^e_i)) = \widehat{\err}_{\hat{T}}(y, f).\label{eq:err_hatt_empirical}
\end{align}
In addition, from the triangle inequality of $\mathcal{L}$ and the assumption $L_0=\sup_{y\in\mathcal{C}} \mathcal{L}(0, y)<\infty$, following inequality holds:
\begin{align}
    0\leq g(Z)&=\mathcal{L}(Y, f(X^c, X^e))\nonumber\\
    &= \mathcal{L}(Y, 0) + \mathcal{L}(Y, f(X^c, X^e)) - \mathcal{L}(Y, 0)\nonumber\\
    &\leq L_0 + k|f(X^c, X^e) - 0|\nonumber\\
    &\leq L_0 + ka\Lambda.\label{eq:gz_bound}
\end{align}
The last inequality follows from the following inequality.
\begin{align*}
    \sup_{\bb x=(\bb x^c, \bb x^e)\in\Omega_t} |f(\bb x)|&=\sup_{\bb x\in\Omega_t} |\langle f, K(\cdot, \bb x)\rangle_\mathcal{H}|\nonumber\\
    &\leq \sup_{\bb x\in\Omega_t} \|f\|_\mathcal{H}\|K(\cdot, \bb x)\|_\mathcal{H}\nonumber\\
    &=\|f\|_\mathcal{H}\sqrt{K(\bb x, \bb x)}\leq a\Lambda,
\end{align*}
where the second line comes from Cauchy-Schwarz inequality, and the last inequality is obtained by the assumption $\sup_{\bb x\in\Omega_t}K(\bb x, \bb x)\leq \Lambda^2$.
By substituting Eqs.~\eqref{eq:err_hatt}, \eqref{eq:err_hatt_empirical}, \eqref{eq:gz_bound} into Eq.~\eqref{eq:uniform_law_of_large_numbers}, we get
\begin{equation}
    \err_{\hat{T}} \leq \widehat{\err}_{\hat{T}} + 2\mathfrak{R}_{N_t} (\mathcal{L}\circ\mathcal{F}) + (L_0+ka\Lambda)\sqrt{\frac{\ln (1/\delta)}{2n}}.
    \label{eq:upperbound_1}
\end{equation}
Then, we can use the Talagrand's contraction lemma~\cite{ledoux2013probability} to derive an upper bound of $\mathfrak{R}_{N_t}(\mathcal{L}\circ\mathcal{F})$.
\begin{lemma}[Talagrand's contraction principle]
Let $\mathcal{G}$ be a set of functions and suppose $\tau:\mathbb{R}\to\mathbb{R}$ is $k$-Lipschitz continuous function.
Then, for any $Z_1, Z_2,\ldots ,Z_n$,
\begin{equation}
    \mathfrak{R}_n(\tau\circ\mathcal{G}) \leq k\mathfrak{R}_n(\mathcal{G}). \label{eq:ub_LF}
\end{equation}
\end{lemma}

Furthermore, when we use $\mathcal{F}=\{f\in\mathcal{H}\mid\|f\|_\mathcal{H}\leq a\}$, we can derive an upper bound of $\mathfrak{R}_{N_t}(\mathcal{F})$ using $N_t$ samples as follows.
Let $\sigma=(\sigma_1, \ldots ,\sigma_{N_t})$ where $\sigma_i\ (i=1,2,\ldots ,N_t)$ are i.i.d and each $\sigma_i$ follows the distribution $P(\sigma_i=+1)=P(\sigma_i=-1)=1/2$.
Then for any fixed $\bb x_1, \bb x_2\ldots ,\bb x_{N_t}$,
\begin{align}
    \mathfrak{R}_{N_t}(\mathcal{F})&=\mathbb{E}_\sigma\left[\sup_{f\in\mathcal{F}}\frac{1}{N_t}\sum_{i=1}^{N_t}\sigma_i f(X_i)\right]
    =\frac{1}{N_t}\mathbb{E}_\sigma\left[\sup_{f\in\mathcal{F}}\sum_{i=1}^n\sigma_i\langle f, K(\cdot, X_i)\rangle\right]\nonumber\\
    &=\frac{1}{N_t}\mathbb{E}_\sigma\left[\sup_{f\in\mathcal{F}}\left\langle f, \sum_{i=1}^{N_t}\sigma_iK(\cdot, X_i)\right\rangle\right]\nonumber\\
    &=\frac{1}{N_t}\mathbb{E}_\sigma\left[\left\langle a\frac{\sum_{i=1}^{N_t}\sigma_i K(\cdot, X_i)}{\|\sum_{i=1}^{N_t}\sigma_i k(\cdot, X_i)\|},\sum_{i=1}^{N_t}\sigma_iK(\cdot, X_i)\right\rangle\right]\label{eq:CS_equal}\\
    &=\frac{a}{N_t}\mathbb{E}_\sigma\left[\left\|\sum_{i=1}^{N_t} \sigma_i K(\cdot, X_i)\right\|\right]
    =\frac{a}{N_t}\mathbb{E}_\sigma\left[\sqrt{\left\|\sum_{i=1}^{N_t} \sigma_i K(\cdot, X_i)\right\|^2}\right]\nonumber\\
    &\leq\frac{a}{N_t}\sqrt{\mathbb{E}_\sigma\left\|\sum_{i=1}^{N_t} \sigma_i K(\cdot, X_i)\right\|^2}\label{eq:Jensen}\\
    &=\frac{a}{N_t}\sqrt{\sum_{i=1}^{N_t} \|K(\cdot, X_i)\|^2}\label{eq:sigma_independence}\\
    &=\frac{a}{N_t}\sqrt{\sum_{i=1}^{N_t} K(X_i, X_i)}
    \leq\frac{a}{N_t}\sqrt{N_t\Lambda^2}\nonumber\\
    &=\frac{a\Lambda}{\sqrt{N_t}}\label{eq:ub_RF}
\end{align}
The equation \eqref{eq:CS_equal} follows from the equality condition of Cauchy-Schwartz inequality, and Jensen's inequality is used to derive Eq.~\eqref{eq:Jensen}.
Equation~\eqref{eq:sigma_independence} follows from $\sigma_i\in\{-1, 1\}$ and the independence of $\sigma_i\ (i=1,\ldots ,N_t)$.

In conclusion, recall the assumption that $\mathcal{L}$ is $k$-Lipchitz function as $|\mathcal{L}(y_1, y_2)-\mathcal{L}(y_1, y_3)|\leq k|y_2-y_3|$ and by applying Eqs.~\eqref{eq:ub_LF}, \eqref{eq:ub_RF} to Eq.~\eqref{eq:upperbound_1} yields, for all $\delta\in (0, 1)$, with probability at least $1-\delta$,
\begin{equation}
    \err_{\hat{T}}(y, \hat{f}_s)\leq \widehat{\err}_{\hat{T}}(y, \hat{f}_s)+\frac{2ka\Lambda}{\sqrt{N_t}} + (L_0+ka\Lambda)\sqrt{\frac{\ln (1/\delta)}{2N_t}}.
    \label{eq:ub_first_term}
\end{equation}

Next, we consider an upper bound of the second term, $\err_{\hat{T}}(y, f_0)$, in Eq.~\eqref{eq:triangular1}.
Recall that the model $f_0$ minimizes the true target error, namely, $f_0 = \argmin_{f\in\mathcal{F}}\err_T(y, f)$.
Here, we consider the absolute difference between $\err_{\hat{T}}(y, f_0)$ and $\err_{T}(y, f_0)$.
Intuitively, when the estimated and true target distributions $\mathcal{P}_{\hat{T}}$ and $\mathcal{P}_T$ are sufficiently close,
the difference between these errors becomes small; hence this difference will be upper bounded.
Here, we consider the upper bound of $|\err_{\hat{T}}(y, f_0)-\err_{T}(y, f_0)|$ with reference to the proof in \cite{NIPS2017_0070d23b} as follows.
\allowdisplaybreaks[2]
\begin{align}
    &|\err_{\hat{T}}(y, f_0)-\err_T(y, f_0)|\nonumber\\
    =&\left|\int_{\Omega_t\times \mathcal{C}}\mathcal{L}(y, f_0(\bb x^c, \bb x^e)) d(\mathcal{P}_{\hat{T}}-\mathcal{P}_T)\right|\nonumber\\
    \leq&\left|\int_{(\Omega_t\times \mathcal{C})^2}\mathcal{L}(\hat{y}_t,f_0(\hat{\bb x}^c_t, \hat{\bb x}^e_t))-\mathcal{L}(y_t, f_0(\bb x^c_t,\bb x^e_t)) d\gamma^*((\hat{\bb x}^c_t, \hat{\bb x}^e_t, \hat{y}_t), (\bb x^c_t, \bb x^e_t,y_t))\right|\label{eq:kantrovich_rubinstein}\\
    \leq &\int_{(\Omega_t\times \mathcal{C})^2}|\mathcal{L}(\hat{y}_t,f_0(\hat{\bb x}^c_t, \hat{\bb x}^e_t))-\mathcal{L}(y_t, f_0(\bb x^c_t, \bb x^e_t))| d\gamma^*((\hat{\bb x}^c_t, \hat{\bb x}^e_t, \hat{y}_t), (\bb x^c_t, \bb x^e_t,y_t))\nonumber\\
    =&\int_{(\Omega_t\times \mathcal{C})^2}|\mathcal{L}(\hat{y}_t,f_0(\hat{\bb x}^c_t, \hat{\bb x}^e_t))-\mathcal{L}(\hat{y}_t,f_0(\bb x^c_t, \bb x^e_t))+\mathcal{L}(\hat{y}_t,f_0(\bb x^c_t, \bb x^e_t))-\mathcal{L}(y_t, f_0(\bb x^c_t, \bb x^e_t))|\nonumber\\&\hspace{8cm}d\gamma^*((\hat{\bb x}^c_t, \hat{\bb x}^e_t, \hat{y}_t), (\bb x^c_t, \bb x^e_t,y_t))\nonumber\\
    \leq&\int_{(\Omega_t\times \mathcal{C})^2}|\mathcal{L}(\hat{y}_t,f_0(\hat{\bb x}^c_t, \hat{\bb x}^e_t))-\mathcal{L}(\hat{y}_t,f_0(\bb x^c_t, \bb x^e_t))|+|\mathcal{L}(\hat{y}_t,f_0(\bb x^c_t, \bb x^e_t))-\mathcal{L}(y_t, f_0(\bb x^c_t, \bb x^e_t))|\nonumber\\&\hspace{8cm} d\gamma^*((\hat{\bb x}^c_t, \hat{\bb x}^e_t, \hat{y}_t), (\bb x^c_t, \bb x^e_t,y_t))\nonumber\\
    \leq &\int_{(\Omega_t\times \mathcal{C})^2}|\mathcal{L}(\hat{y}_t,f_0(\hat{\bb x}^c_t, \hat{\bb x}^e_t))-\mathcal{L}(\hat{y}_t,f_0(\bb x^c_t, \bb x^e_t))| + \mathcal{L}(\hat{y}_t,y_t) d\gamma^*((\hat{\bb x}^c_t, \hat{\bb x}^e_t, \hat{y}_t), (\bb x^c_t, \bb x^e_t,y_t))\label{eq:ell_triangular}\\
    \leq &\int_{(\Omega_t\times \mathcal{C})^2}k|f_0(\hat{\bb x}^c_t, \hat{\bb x}^e_t)-f_0(\bb x^c_t,\bb x^e_t)| + \mathcal{L}(\hat{y}_t,y_t) d\gamma^*((\hat{\bb x}^c_t, \hat{\bb x}^e_t, \hat{y}_t), (\bb x^c_t, \bb x^e_t,y_t))\label{eq:loss_lipschitzness}\\
    \leq &\int_{(\Omega_t\times \mathcal{C})^2}k\lambda d((\hat{\bb x}^c_t, \hat{\bb x}^e_t),(\bb x^c_t,\bb x^e_t)) + \mathcal{L}(\hat{y}_t,y_t) d\gamma^*((\hat{\bb x}^c_t, \hat{\bb x}^e_t, \hat{y}_t), (\bb x^c_t, \bb x^e_t,y_t))+kM\phi(\lambda)\label{eq:f0_PTL}\\
    = & \int_{(\Omega_t\times \mathcal{C})^2}\alpha d((\hat{\bb x}^c_t, \hat{\bb x}^e_t),(\bb x^c_t,\bb x^e_t)) + \mathcal{L}(\hat{y}_t,y_t) d\gamma^*((\hat{\bb x}^c_t, \hat{\bb x}^e_t, \hat{y}_t), (\bb x^c_t, \bb x^e_t,y_t))+kM\phi(\lambda)\nonumber\\
    = &W(\mathcal{P}_{\hat{T}},\mathcal{P}_T) + kM\phi(\lambda)\label{eq:ub_second_term}
\end{align}
where $\alpha=k\lambda$. 
The line \eqref{eq:kantrovich_rubinstein} is a consequence of the duality form of the Kantrovich-Rubinstein theorem.
The line~\eqref{eq:ell_triangular} follows from the assumption that $\mathcal{L}$ satisfies the triangular inequality.
The line~\eqref{eq:loss_lipschitzness} comes from the $k$-Lipschitzness of the loss function $\mathcal{L}(\cdot, y)$.
Finally, the line~\eqref{eq:f0_PTL} follows from the probabilistic transfer Lipschitzness of $f_0$ and $\gamma^*$.
More specifically, with probability at least $1-\phi(\lambda)$, the model $f_0$ verify the $\lambda$-Lipschitzness with distance function $d$, then
\begin{equation*}
    |f_0(\hat{\bb x}^c_t, \hat{\bb x}^e_t)-f_0(\bb x^c_t,\bb x^e_t)|\leq \lambda d((\hat{\bb x}^c_t, \hat{\bb x}^e_t), (\bb x^c_t,\bb x^e_t))
\end{equation*}
On the other hand, with probability at most $\phi(\lambda)$, 
\begin{equation*}
    |f_0(\hat{\bb x}^c_t, \hat{\bb x}^e_t)-f_0(\bb x^c_t,\bb x^e_t)|> \lambda d((\hat{\bb x}^c_t, \hat{\bb x}^e_t), (\bb x^c_t,\bb x^e_t))
\end{equation*}
However, in this case,
for all $\hat{\bb x}^c_t, \hat{\bb x}^e_t,\bb x^c_t$, and $\bb x^e_t$,
\begin{equation*}
    |f_0 (\hat{\bb x}^c_t, \hat{\bb x}^e_t) - f_0(\bb x^c_t, \bb x^e_t)|\leq M
\end{equation*}
holds from the assumption.
The line \eqref{eq:f0_PTL} is derived using these inequalities.

From Eq.~\eqref{eq:ub_second_term},
\begin{equation}
    \err_{\hat{T}}(y, f_0)\leq \err_T(y, f_0)+W(\mathcal{P}_{\hat{T}}, \mathcal P_T) + kM\phi(\lambda) \label{eq:ub_second_term2}
\end{equation}
holds.
Substituting Eqs.~\eqref{eq:ub_first_term}, \eqref{eq:ub_second_term2} into Eq.~\eqref{eq:triangular1} yields Theorem 1.
\section{The Effect of the Choice of Parameter and Model Set}

\subsection{The Effect of the Balancing Parameter}
\begin{figure}[ht]
    \centering
    \begin{tabular}{c}
    \begin{minipage}[b]{0.35\linewidth}
        \centering
        \includegraphics[width=\textwidth]{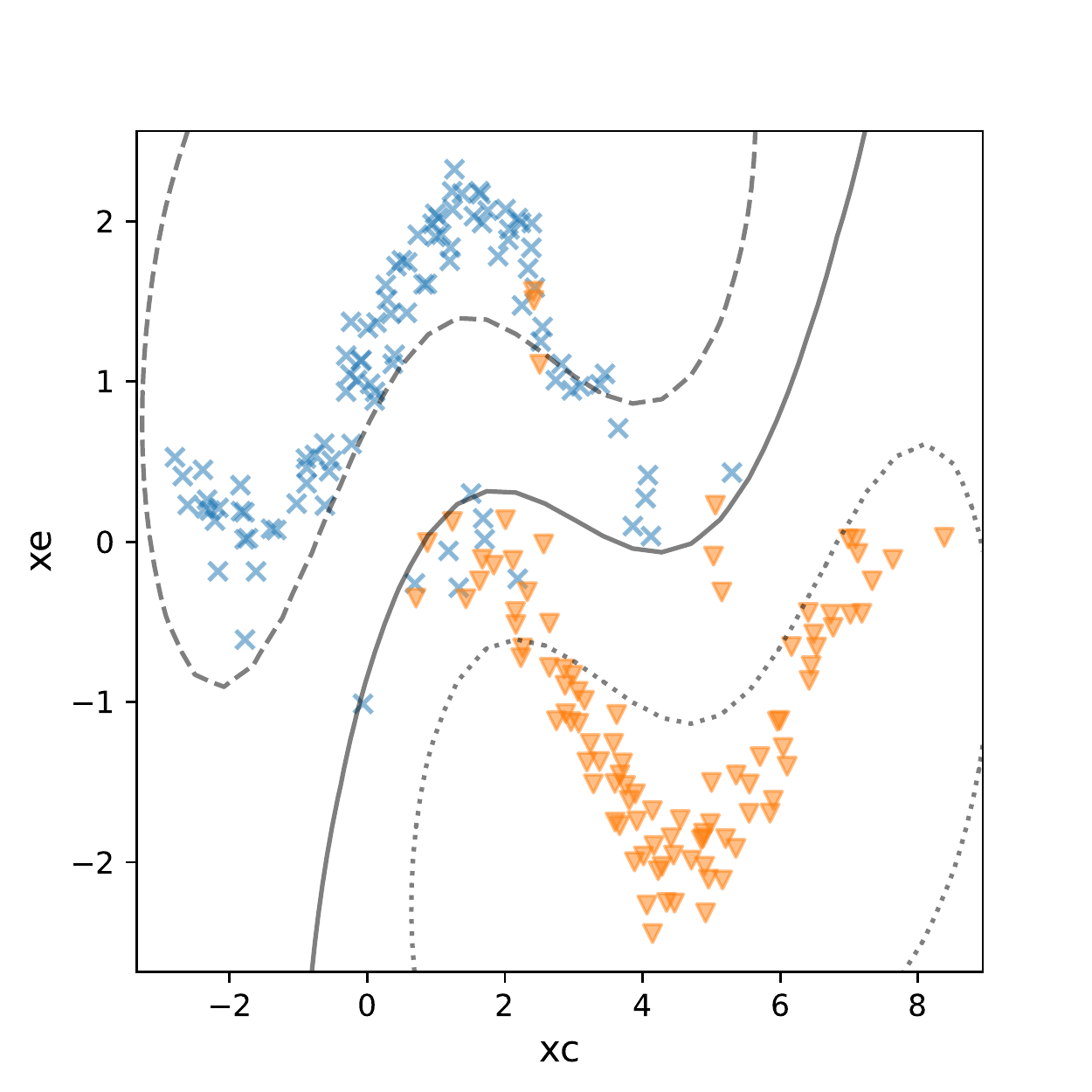}
        \subcaption{$\alpha=100$}\label{fig:alpha_0.01}
    \end{minipage}
      \begin{minipage}[b]{0.35\linewidth}
        \centering
        \includegraphics[width=\textwidth]{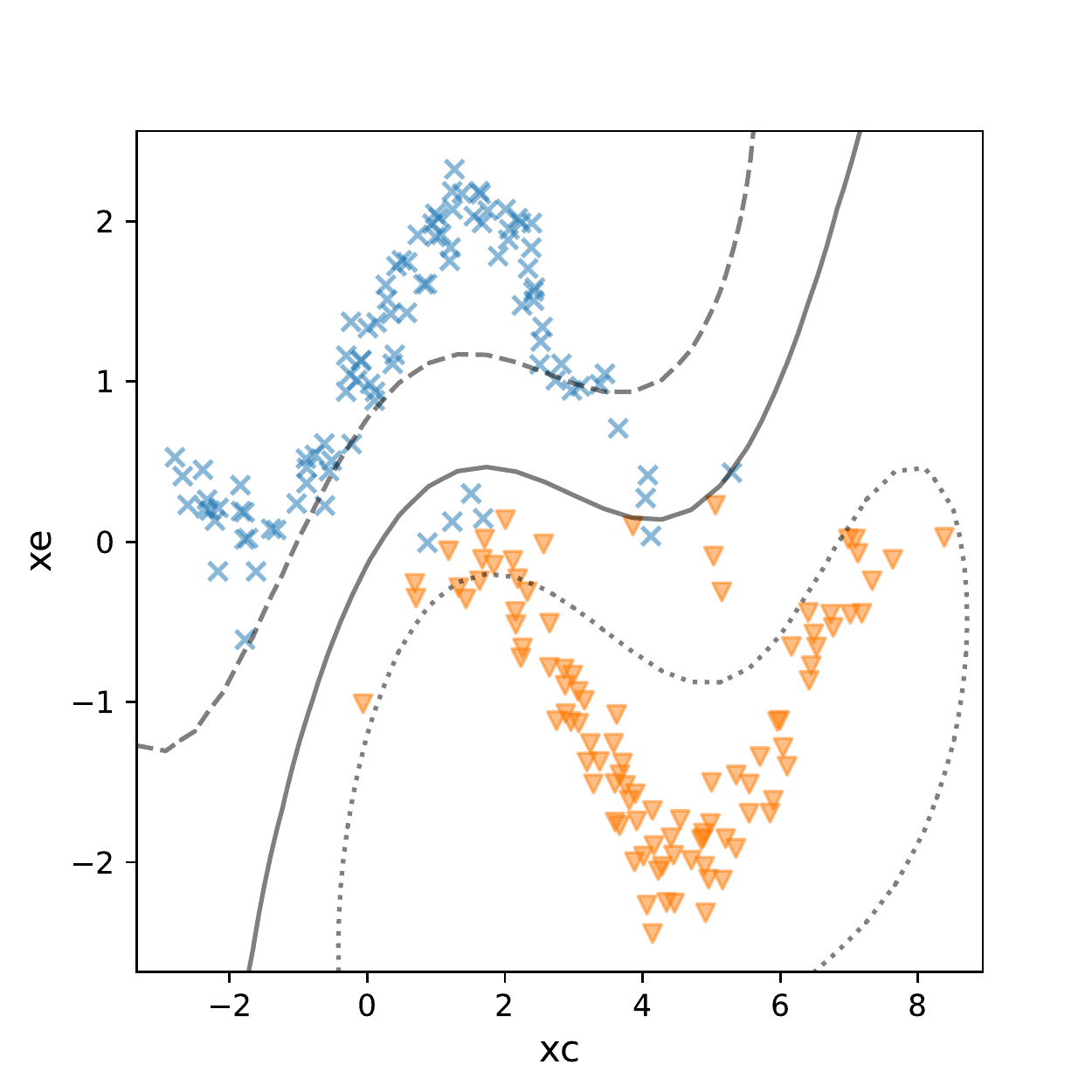}
        \subcaption{$\alpha=10$}\label{fig:alpha_0.1}
      \end{minipage}\\
          \begin{minipage}[b]{0.35\linewidth}
        \centering
        \includegraphics[width=\textwidth]{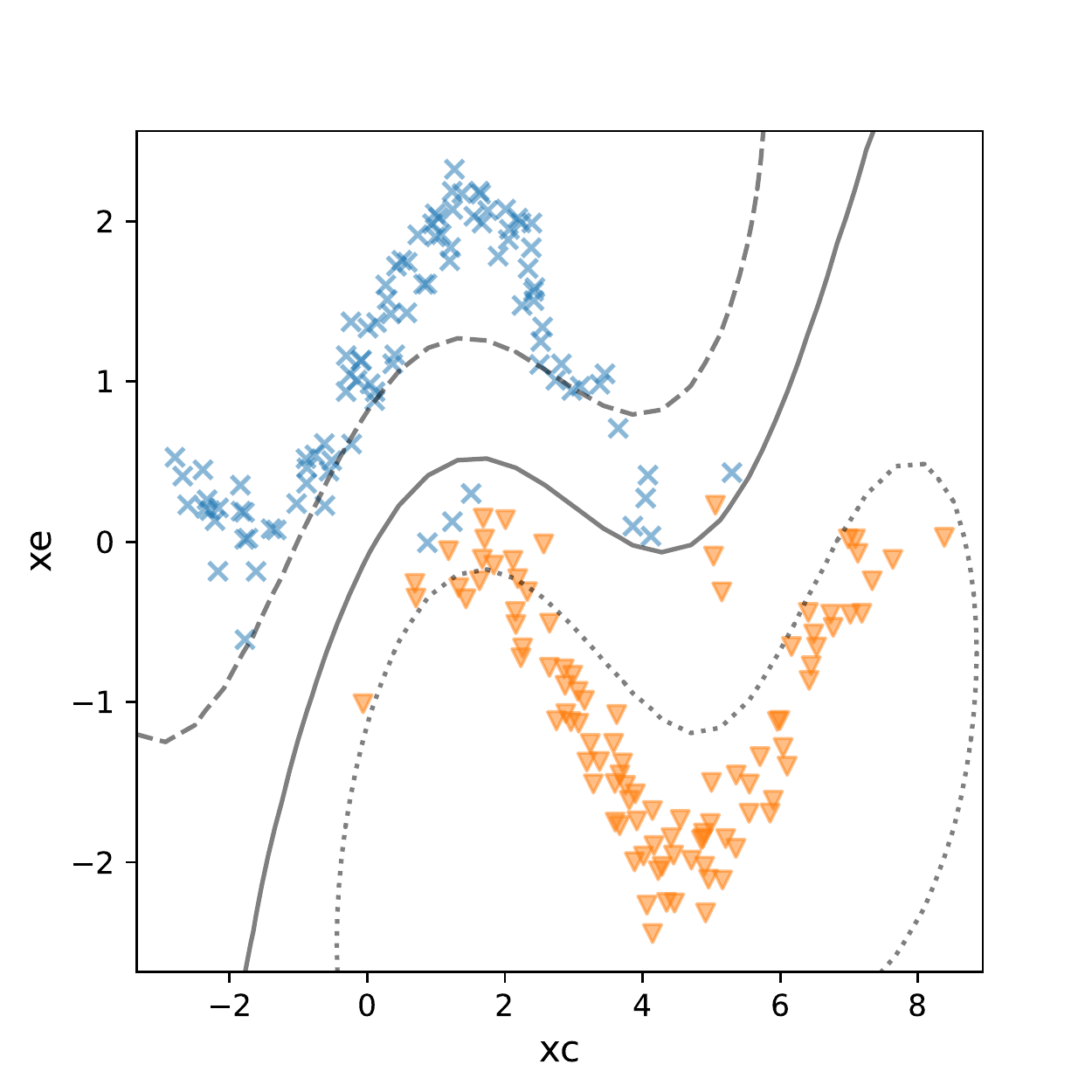}
        \subcaption{$\alpha=1.0$}\label{fig:alpha_1}
    \end{minipage}
      \begin{minipage}[b]{0.35\linewidth}
        \centering
        \includegraphics[width=\textwidth]{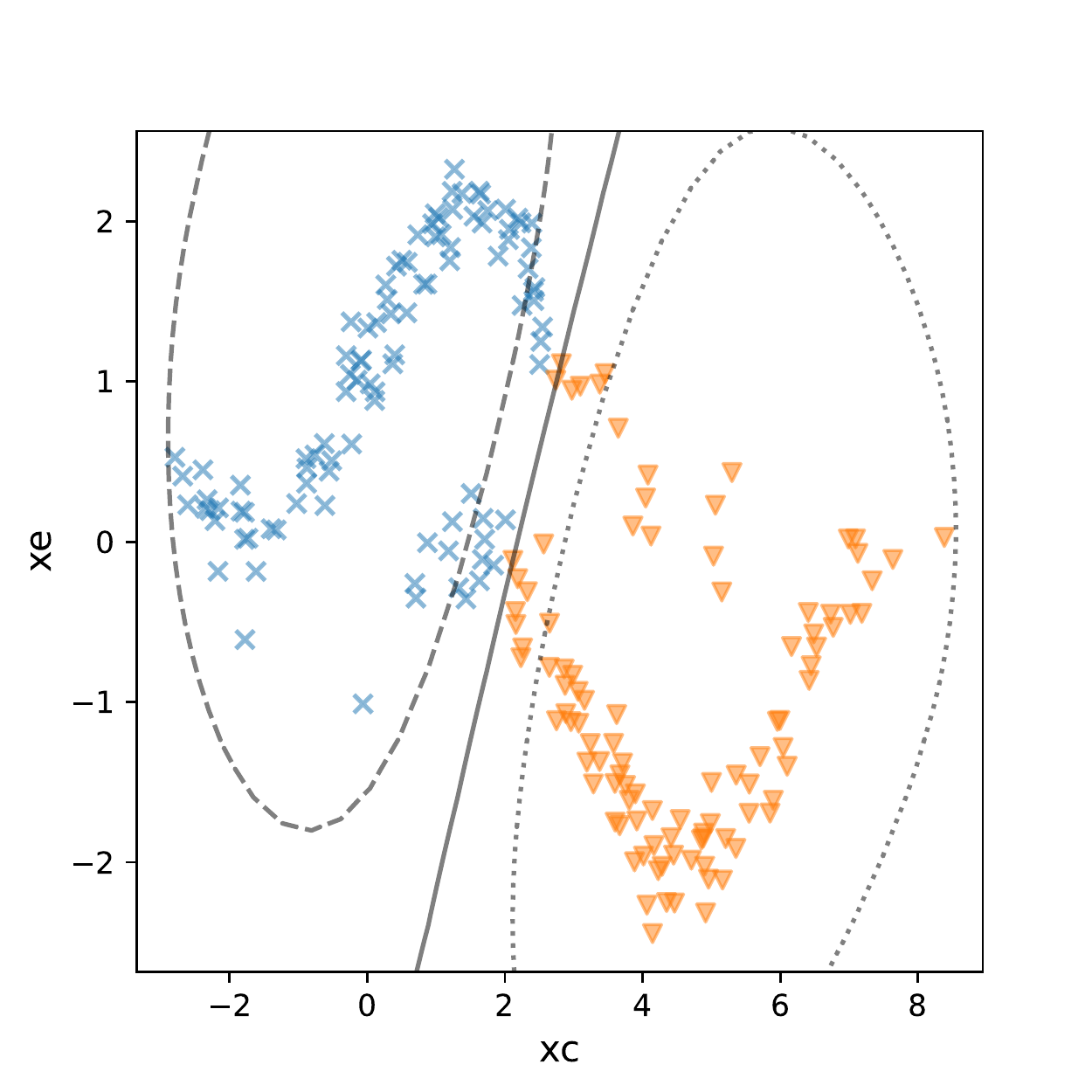}
        \subcaption{$\alpha=0.10$}\label{fig:alpha_10}
      \end{minipage}\\
      \end{tabular}
      \caption{Obtained decision boundary for various $\alpha$. The solid line shows the decision boundary $f(x^c, x^e)=0$ and the dashed lines shows the margins where $f(x^c, x^e)=\pm 1$, respectively.}\label{fig:result_supp}
\end{figure}

We determined the value of the balancing parameter as $\alpha=1.0$ by preliminary experiments in the main manuscript.
The value of $\alpha$ affects the optimal solution of the problem Eq.~(3) and also the training process of the model $f$ in Eq.~(6).
In this subsection, we experimentally assess the effect of the value of the balancing parameter $\alpha$ in the transportation cost
\begin{equation}
    \mathcal{E}_\alpha (\bb x^c_s, y_s; \bb x^c_t, \bb x^e_t, y_t)\equiv \alpha d(\bb x^c_s,\bb x^c_t)+\mathcal{L}(y_s, y_t).
\end{equation}
The dataset used in this experiment is the same as in Figure~\ref{fig:dataset_example}.
Note that the extra feature of the source samples and the label of the target samples are not available in the experiment as stated in the main text.
Figure \ref{fig:result_supp} shows the experimental results for $\alpha=100, 10, 1.0, 0.10$ with the same dataset.
As we can see from the figure, the value of $\alpha$ does not have large impact, however, for too small $\alpha$, the error between the transferred labels and the pseudo-labels becomes dominant; hence, the source samples are transferred so that they are fitted to the temporal model.
The results suggest that when the value of $\alpha$ is too small, both the training of a model and the OT tend to be stuck in the local minima.
To develop a method to determine an appropriate value for $\alpha$ is left for our future work.

\subsection{The Effect of the Choice of a Model Set}
\begin{figure}[b!]
    \centering
    \begin{minipage}[b]{0.30\linewidth}
        \centering
        \includegraphics[width=\textwidth]{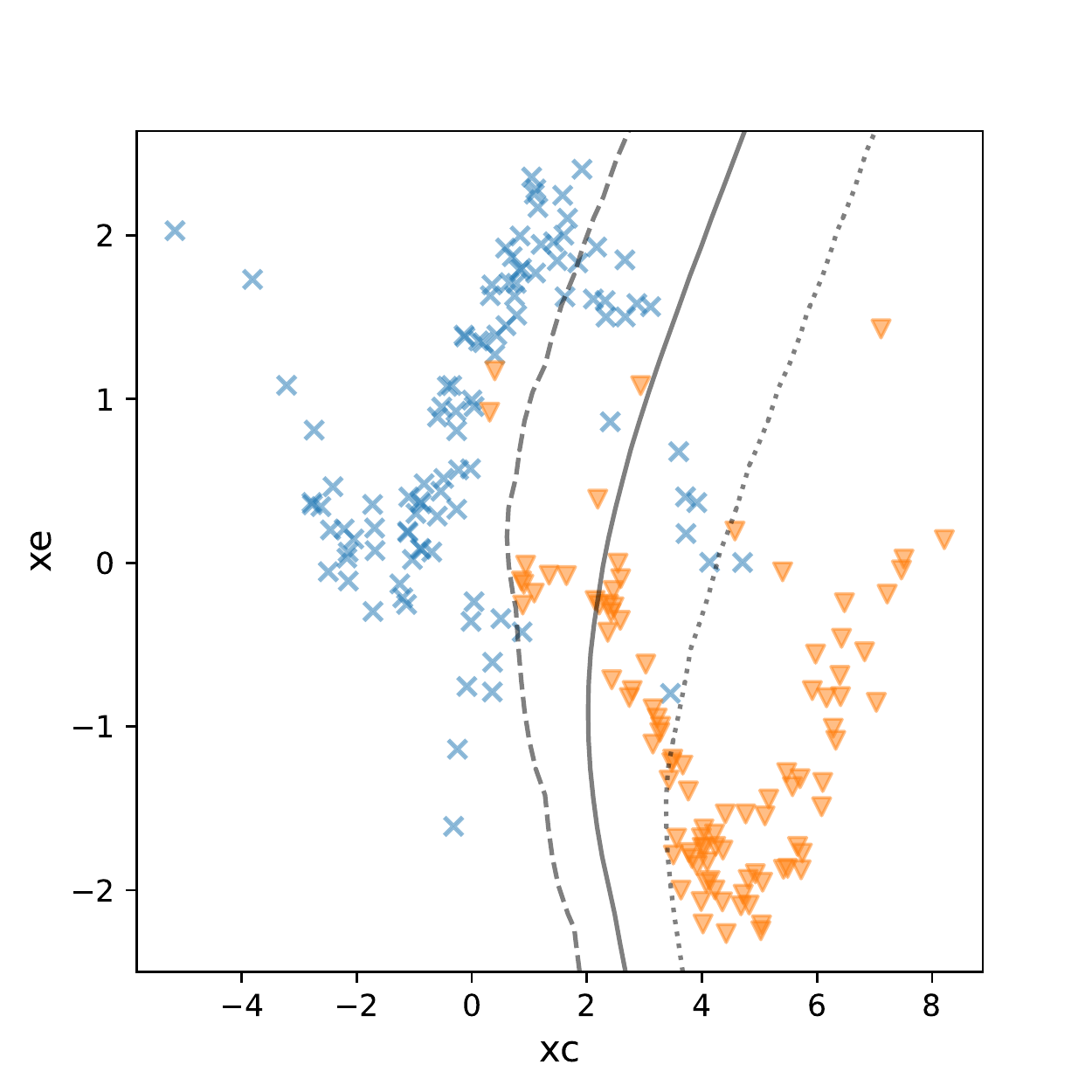}
        \subcaption{iteration 1}
    \end{minipage}
      \begin{minipage}[b]{0.30\linewidth}
        \centering
        \includegraphics[width=\textwidth]{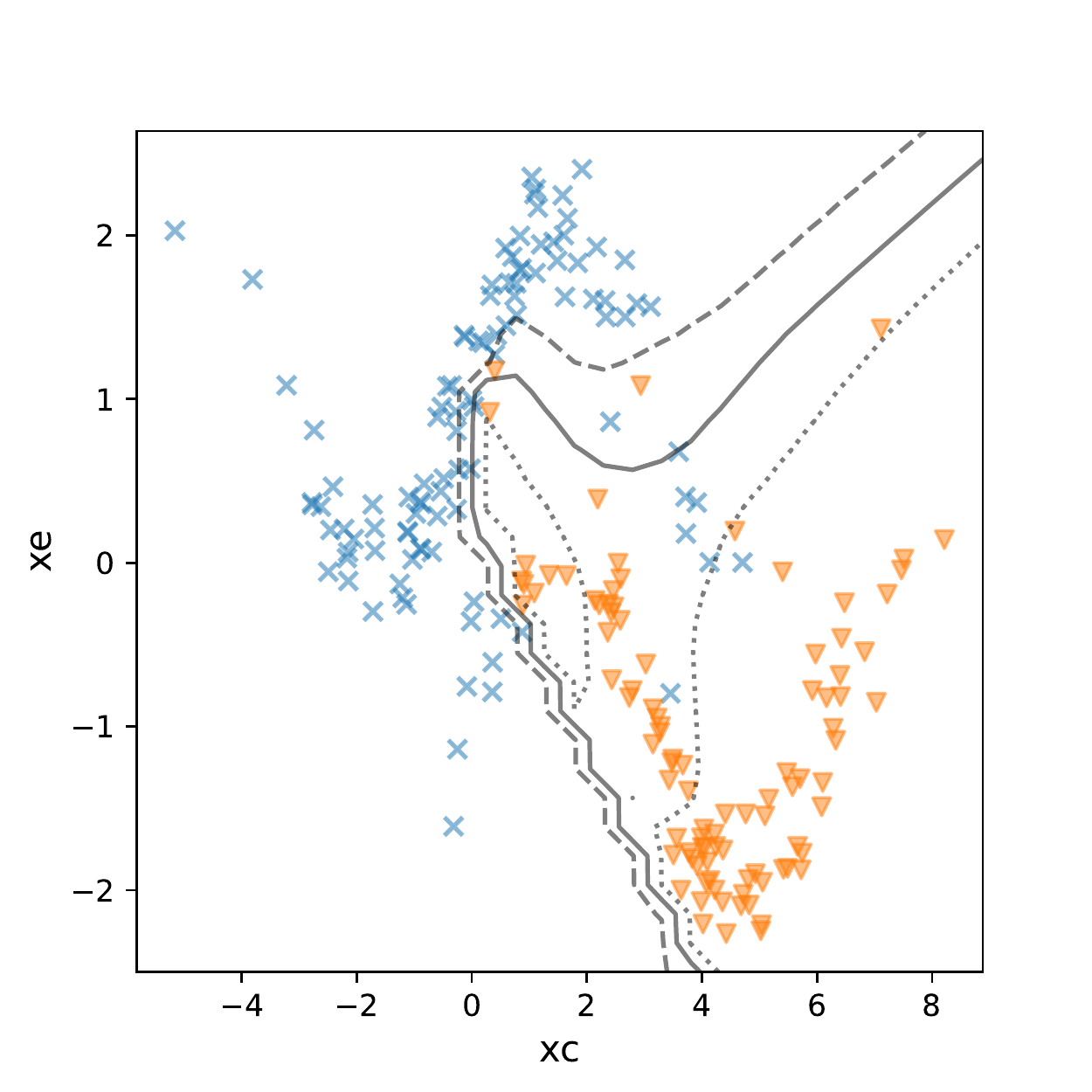}
        \subcaption{iteration 10}
      \end{minipage}
      \begin{minipage}[b]{0.30\linewidth}
        \centering
        \includegraphics[width=\textwidth]{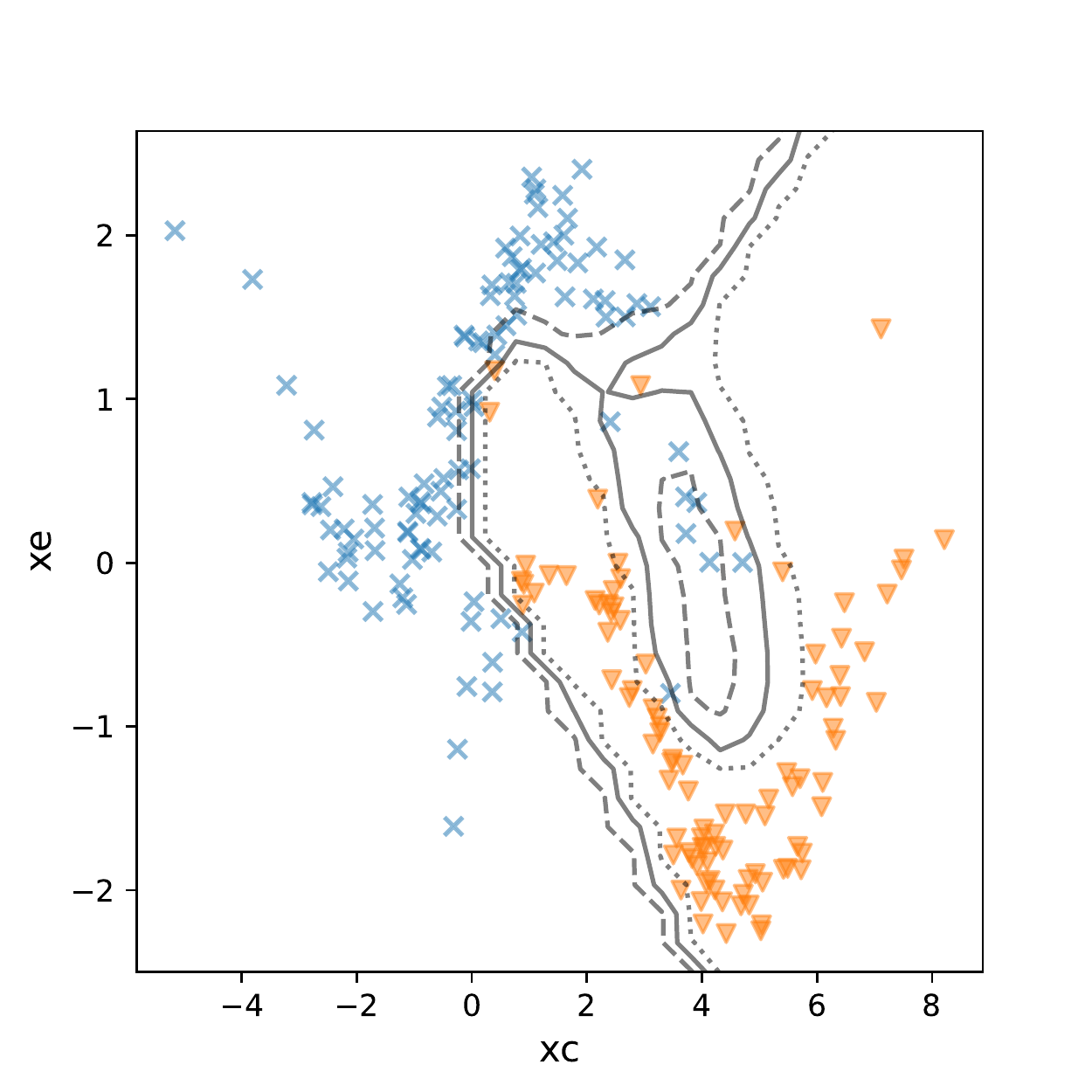}
        \subcaption{iteration 20}
      \end{minipage}
    \caption{Decision boundary of trained three-layer neural network for each iteration. The model estimates the positive probability of instances and $f(x^c, x^e)=0.5$ is shown by the solid line, and the curves of $f(x^c, x^e)=0.05, 0.95$ are shown by dashed and dotted lines respectively.} 
    \label{fig:boundary_nn}
\end{figure}

In the experiment in the manuscript, we used a set of support vector machines as a model set $\mathcal{F}$.
Here, we show experimental results where the model set is a set of four-layer neural networks where the numbers of the hidden neurons are 512, and 64.
We used the same dataset as the previous subsection for the experiment, then Figure \ref{fig:boundary_nn} shows the decision boundary of the trained model for the first 20 iterations.
As we can see from the figure, the transferred labels do not estimate the true labels accurately,
and the trained model is fitted to the inaccurate target labels.
In our proposed method, the cost of the OT in the first iteration considers only the distance between the common features of the source and target domains; therefore, the transferred labels are inaccurate in the first iteration.
Once the model is fitted to the inaccurate labels, subsequent estimation of the target labels by OT also becomes inaccurate because the pseudo-labels produced by the fitted model are used to calculate the transportation cost.
In general, complex models can be fitted to the initial inaccurate labels; hence, the prediction of the target labels by the complex model becomes less accurate. To use the proposed method with highly flexible models, the complexity should be controlled by appropriate regularization or by the early stopping of the learning process.

\section{Additional Experiments}
\subsection{Additional Experiments with Gas Sensor Array Data}
In this subsection, we show the experimental results obtained with the gas sensor array dataset.
Only the first 4 batches are used for the experiment in the main manuscript,
here we show the model accuracy in the test domain for all the pairs of domains.
We consider the binary classification problem to classify ethanol and ethylene in the same manner as the experiments in the main manuscript.
We chose 8 out of 16 sensors and calculate the accuracy of the trained model in the target domain.

We remark that our proposed method implicitly assumes that the class balance is consistent across the source and target domains.
Although this assumption does not hold for real data and the balance of the class differs for each domain, we applied our proposed method to real data without modifying the class balance.

Table~\ref{tab:all_res_gas} shows the experimental results with all pairs of domains.
As we can see from the table, domain adaptation is not required between some domains, and the proposed method does not necessarily work well for all domains.
In particular, for the latter half of domains (domain 6, 7, 8, 9, and 10) baseline accuracy tend to outperform domain adaptation methods.
Since the domain are arranged in chronological order, this implies that the appropriate sensors that can be used for the common feature for the domain adaptation change in time.
When the appropriate sensors for the common features are changed, the estimation of the target distribution by OT using the predefined sensors becomes difficult.
Therefore, the baseline model outperforms both JDOT and the proposed method in such cases.
\begin{table}[t]
\renewcommand{\arraystretch}{0.9}
\scriptsize
\centering
\caption{Accuracy of target model with gas sensor array dataset. Each line of cells shows baseline accuracy, accuracy by JDOT without extra features, CCA-based heterogeneous domain adaptation, and domain specific feature transfer, and the proposed method.}
\begin{tabular}{llllllllllll}
\cline{2-12}
\multicolumn{1}{l|}{}&&\multicolumn{10}{c|}{target domain}\\\hline
\multicolumn{1}{|l|}{}&\multicolumn{1}{c|}{}&\multicolumn{1}{c|}{1}&\multicolumn{1}{c|}{2}&\multicolumn{1}{c|}{3}&\multicolumn{1}{c|}{4}&\multicolumn{1}{c|}{5}&\multicolumn{1}{c|}{6}&\multicolumn{1}{c|}{7}& \multicolumn{1}{c|}{8} &\multicolumn{1}{c|}{9}&\multicolumn{1}{c|}{10}\\\cline{2-12}
\multicolumn{1}{|l|}{\multirow{30}{*}{\begin{sideways}source domain\end{sideways}}}&
\multicolumn{1}{l|}{1}&
\multicolumn{1}{l|}{}&
\multicolumn{1}{l|}{\begin{tabular}[c]{@{}c@{}}\textbf{83.13}\\ 77.71\\ 66.67\\ 42.97\\ 78.31\end{tabular}}&
\multicolumn{1}{l|}{\begin{tabular}[c]{@{}c@{}}52.51\\ 93.45\\ 54.15\\ 57.78\\ \textbf{96.02}\end{tabular}}&
\multicolumn{1}{l|}{\begin{tabular}[c]{@{}c@{}}64.49\\ 60.75\\43.93\\ \textbf{94.39}\\87.65\end{tabular}}&
\multicolumn{1}{l|}{\begin{tabular}[c]{@{}c@{}}38.24\\ 97.06\\ 69.11\\\textbf{100.0}\\\textbf{100.0}\end{tabular}}&
\multicolumn{1}{l|}{\begin{tabular}[c]{@{}c@{}}61.40\\ 88.71\\ 52.48\\50.09\\\textbf{90.44}\end{tabular}}&
\multicolumn{1}{l|}{\begin{tabular}[c]{@{}c@{}}46.07\\ 89.86\\ 45.16\\49.50\\\textbf{90.19}\end{tabular}}&
\multicolumn{1}{l|}{\begin{tabular}[c]{@{}c@{}}\textbf{50.00}\\ 1.66\\ 50.00\\50.00\\10.00\end{tabular}}&
\multicolumn{1}{l|}{\begin{tabular}[c]{@{}c@{}}52.59\\ 8.621\\\textbf{100.0}\\8.621\\8.621\end{tabular}}&
\multicolumn{1}{l|}{\begin{tabular}[c]{@{}c@{}}77.92\\ \textbf{95.00}\\ 51.83\\ 50.17\\92.42\end{tabular}}\\\cline{2-12}
\multicolumn{1}{|l|}{}&
\multicolumn{1}{l|}{2}&
\multicolumn{1}{l|}{\begin{tabular}[c]{@{}c@{}}52.13\\ 79.26\\48.94\\62.77\\\textbf{84.04}\end{tabular}}&
\multicolumn{1}{l|}{}&
\multicolumn{1}{l|}{\begin{tabular}[c]{@{}c@{}}57.31\\ 89.36\\ 42.81\\84.56\\\textbf{89.47}\end{tabular}}&
\multicolumn{1}{l|}{\begin{tabular}[c]{@{}c@{}}63.55\\ 69.16\\ 40.19\\41.12\\\textbf{71.96}\end{tabular}}&
\multicolumn{1}{l|}{\begin{tabular}[c]{@{}c@{}}60.29\\ 86.76\\\textbf{92.65}\\58.82\\ \textbf{92.65}\end{tabular}}&
\multicolumn{1}{l|}{\begin{tabular}[c]{@{}c@{}}52.76\\ 76.19\\ 51.01\\52.76\\\textbf{76.47}\end{tabular}}&
\multicolumn{1}{l|}{\begin{tabular}[c]{@{}c@{}}50.50\\ 75.44\\ 42.56\\50.50\\\textbf{75.59}\end{tabular}}&
\multicolumn{1}{l|}{\begin{tabular}[c]{@{}c@{}}\textbf{50.00}\\ 11.67\\ \textbf{50.00}\\\textbf{50.00}\\ 11.67\end{tabular}}&
\multicolumn{1}{l|}{\begin{tabular}[c]{@{}c@{}}47.41\\ \textbf{56.90}\\ 47.41\\47.41\\43.97\end{tabular}}&
\multicolumn{1}{l|}{\begin{tabular}[c]{@{}c@{}}47.42\\ \textbf{81.08}\\ 50.00\\50.00\\ 79.42\end{tabular}}\\\cline{2-12}
\multicolumn{1}{|l|}{}&\multicolumn{1}{l|}{3}&
\multicolumn{1}{l|}{\begin{tabular}[c]{@{}c@{}}52.13\\ 92.02\\ 52.66\\66.49\\\textbf{94.15}\end{tabular}}&
\multicolumn{1}{l|}{\begin{tabular}[c]{@{}c@{}}72.09\\ 81.93\\32.73\\ 86.94\\\textbf{88.76}\end{tabular}}&
\multicolumn{1}{l|}{}&
\multicolumn{1}{l|}{\begin{tabular}[c]{@{}c@{}}\textbf{94.39}\\ 77.57\\ 47.66\\40.19\\81.31\end{tabular}}&
\multicolumn{1}{l|}{\begin{tabular}[c]{@{}c@{}}\textbf{100.0}\\ \textbf{100.0}\\ 45.59\\58.82\\\textbf{100.0}\end{tabular}}&
\multicolumn{1}{l|}{\begin{tabular}[c]{@{}c@{}}\textbf{90.35}\\ 83.91\\ 14.89\\76.01\\ 85.66\end{tabular}}&
\multicolumn{1}{l|}{\begin{tabular}[c]{@{}c@{}}79.25\\ \textbf{85.51}\\ 31.81\\55.15\\ 85.35\end{tabular}}&
\multicolumn{1}{l|}{\begin{tabular}[c]{@{}c@{}}50.00\\ 1.67\\ \textbf{53.33}\\50.00\\1.67\end{tabular}}&
\multicolumn{1}{l|}{\begin{tabular}[c]{@{}c@{}}\textbf{52.59}\\ 0.8621\\ 51.72\\ 47.41\\ 0.8621\end{tabular}}&
\multicolumn{1}{l|}{\begin{tabular}[c]{@{}c@{}}50.00\\ 82.33\\46.83\\ 59.42\\ \textbf{85.17}\end{tabular}}\\\cline{2-12}
\multicolumn{1}{|l|}{}&\multicolumn{1}{l|}{4}&
\multicolumn{1}{l|}{\begin{tabular}[c]{@{}c@{}}49.47\\ 52.66\\46.80\\ 51.60\\\textbf{53.72}\end{tabular}}&
\multicolumn{1}{l|}{\begin{tabular}[c]{@{}c@{}}42.97\\ 71.08\\ 66.27\\32.93\\ \textbf{74.30}\end{tabular}}&
\multicolumn{1}{l|}{\begin{tabular}[c]{@{}c@{}}91.11\\ \textbf{82.81}\\80.47\\ 42.69\\\textbf{82.57}\end{tabular}}&
\multicolumn{1}{l|}{}&
\multicolumn{1}{l|}{\begin{tabular}[c]{@{}c@{}}\textbf{100.0}\\ 70.59\\ 17.65\\95.59\\ 70.59\end{tabular}}&
\multicolumn{1}{l|}{\begin{tabular}[c]{@{}c@{}}72.61\\ 85.39\\ \textbf{93.75}\\47.24\\85.48\end{tabular}}&
\multicolumn{1}{l|}{\begin{tabular}[c]{@{}c@{}}\textbf{76.51}\\ 68.73\\  48.44\\ 49.50\\ 69.95\end{tabular}}&
\multicolumn{1}{l|}{\begin{tabular}[c]{@{}c@{}}33.33\\ 0\\\textbf{53.33}\\0.00\\ 8.33 \end{tabular}}&
\multicolumn{1}{l|}{\begin{tabular}[c]{@{}c@{}}52.59\\3.45\\\textbf{53.45}\\52.59\\3.45\end{tabular}}&
\multicolumn{1}{l|}{\begin{tabular}[c]{@{}c@{}}50.00\\ \textbf{54.42}\\47.00\\ 52.59\\ 50.00\end{tabular}}\\\cline{2-12}
\multicolumn{1}{|l|}{}&\multicolumn{1}{l|}{5}&
\multicolumn{1}{l|}{\begin{tabular}[c]{@{}c@{}}47.34\\ 91.49\\53.19\\ 62.77\\\textbf{92.02}\end{tabular}}&
\multicolumn{1}{l|}{\begin{tabular}[c]{@{}c@{}}41.16\\ 85.94\\ 67.07\\67.06\\\textbf{89.56}\end{tabular}}&
\multicolumn{1}{l|}{\begin{tabular}[c]{@{}c@{}}\textbf{91.46}\\ 91.11\\ 66.08\\57.31\\ 91.11\end{tabular}}&
\multicolumn{1}{l|}{\begin{tabular}[c]{@{}c@{}}76.64\\ 74.76\\ 41.12\\\textbf{94.39}\\77.57\end{tabular}}&
\multicolumn{1}{l|}{}&
\multicolumn{1}{l|}{\begin{tabular}[c]{@{}c@{}}60.94\\ \textbf{94.76}\\ 46.05\\52.76\\89.61\end{tabular}}&
\multicolumn{1}{l|}{\begin{tabular}[c]{@{}c@{}}67.12\\ \textbf{92.14}\\ 15.03\\50.50\\91.99\end{tabular}}&
\multicolumn{1}{l|}{\begin{tabular}[c]{@{}c@{}}0\\ 78.33\\ \textbf{98.33}\\48.33\\73.33\end{tabular}}&
\multicolumn{1}{l|}{\begin{tabular}[c]{@{}c@{}}0\\ \textbf{50.00}\\ 49.14\\ 47.41\\43.10\end{tabular}}&
\multicolumn{1}{l|}{\begin{tabular}[c]{@{}c@{}}36.92\\ 64.17\\40.75\\50.00\\ \textbf{67.42}\end{tabular}}\\\cline{2-12}
\multicolumn{1}{|l|}{}&\multicolumn{1}{l|}{6}&
\multicolumn{1}{l|}{\begin{tabular}[c]{@{}c@{}}75.00\\ 94.68\\ 50.00\\54.79\\\textbf{98.40}\end{tabular}}&
\multicolumn{1}{l|}{\begin{tabular}[c]{@{}c@{}}\textbf{87.35}\\ 86.14\\ 67.07\\ 69.68\\86.55\end{tabular}}&
\multicolumn{1}{l|}{\begin{tabular}[c]{@{}c@{}}90.64\\ 89.71\\ 28.42\\62.81\\ \textbf{93.22}\end{tabular}}&
\multicolumn{1}{l|}{\begin{tabular}[c]{@{}c@{}}74.77\\ \textbf{81.31}\\40.19\\ 40.19\\80.37\end{tabular}}&
\multicolumn{1}{l|}{\begin{tabular}[c]{@{}c@{}}\textbf{100.0}\\ \textbf{100.0}\\58.82\\ 58.82\\ \textbf{100.0}\end{tabular}}&
\multicolumn{1}{l|}{}&
\multicolumn{1}{l|}{\begin{tabular}[c]{@{}c@{}}\textbf{100.0}\\ 99.69\\ 43.78\\88.33\\ 97.33\end{tabular}}&
\multicolumn{1}{l|}{\begin{tabular}[c]{@{}c@{}}\textbf{100.0}\\ \textbf{100.0}\\ 50.00\\ 50.00\\ 98.33\end{tabular}}&
\multicolumn{1}{l|}{\begin{tabular}[c]{@{}c@{}}\textbf{100.0}\\ \textbf{100.0}\\ \textbf{100.0}\\47.41\\ 99.14\end{tabular}}&
\multicolumn{1}{l|}{\begin{tabular}[c]{@{}c@{}}\textbf{96.42}\\ 92.75\\ 63.00\\85.67\\ 89.50\end{tabular}}\\\cline{2-12}
\multicolumn{1}{|l|}{}&\multicolumn{1}{l|}{7}&
\multicolumn{1}{l|}{\begin{tabular}[c]{@{}c@{}}59.57\\ 91.49\\61.70\\54.79\\ \textbf{94.15}\end{tabular}}&
\multicolumn{1}{l|}{\begin{tabular}[c]{@{}c@{}}84.34\\ 76.51\\ 33.13\\70.28\\\textbf{84.94}\end{tabular}}&
\multicolumn{1}{l|}{\begin{tabular}[c]{@{}c@{}}86.43\\ 87.13\\ 57.89\\61.64\\\textbf{89.00}\end{tabular}}&
\multicolumn{1}{l|}{\begin{tabular}[c]{@{}c@{}}\textbf{71.96}\\ 65.55\\ 44.86\\40.19\\70.09\end{tabular}}&
\multicolumn{1}{l|}{\begin{tabular}[c]{@{}c@{}}72.06\\ \textbf{100.0}\\ 75.00\\58.52\\\textbf{100.0}\end{tabular}}&
\multicolumn{1}{l|}{\begin{tabular}[c]{@{}c@{}}87.32\\ 98.62\\ 67.10\\75.28\\\textbf{100.0}\end{tabular}}&
\multicolumn{1}{l|}{}&
\multicolumn{1}{l|}{\begin{tabular}[c]{@{}c@{}}\textbf{100.0}\\ 0\\ 50.00\\50.00\\ 0\end{tabular}}&
\multicolumn{1}{l|}{\begin{tabular}[c]{@{}c@{}}\textbf{100.0}\\ 1.72\\ 0.00\\ 47.41\\1.72\end{tabular}}&
\multicolumn{1}{l|}{\begin{tabular}[c]{@{}c@{}}94.92\\ 91.17\\ 46.17\\ \textbf{99.00}\\ 91.33\end{tabular}}\\\cline{2-12}
\multicolumn{1}{|l|}{}&\multicolumn{1}{l|}{8}&
\multicolumn{1}{l|}{\begin{tabular}[c]{@{}c@{}}\textbf{52.13}\\ 45.15\\ \textbf{52.13}\\\textbf{52.13}\\45.21\end{tabular}}&
\multicolumn{1}{l|}{\begin{tabular}[c]{@{}c@{}}58.63\\ 36.34\\ \textbf{67.47}\\ 67.07\\ 36.14\end{tabular}}&
\multicolumn{1}{l|}{\begin{tabular}[c]{@{}c@{}}57.19\\ 33.10\\ 9.591\\ \textbf{57.31}\\ 33.22\end{tabular}}&
\multicolumn{1}{l|}{\begin{tabular}[c]{@{}c@{}}39.25\\ 38.32\\ 14.02\\ \textbf{40.19}\\ 34.58\end{tabular}}&
\multicolumn{1}{l|}{\begin{tabular}[c]{@{}c@{}}42.65\\ \textbf{51.47}\\ 47.06\\32.35\\ \textbf{51.47}\end{tabular}}&
\multicolumn{1}{l|}{\begin{tabular}[c]{@{}c@{}}53.49\\ \textbf{62.68}\\ 44.39\\ 52.76\\62.31\end{tabular}}&
\multicolumn{1}{l|}{\begin{tabular}[c]{@{}c@{}}\textbf{63.92}\\ 39.66\\ 38.83\\ 50.50\\39.36\end{tabular}}&
\multicolumn{1}{l|}{}&
\multicolumn{1}{l|}{\begin{tabular}[c]{@{}c@{}}\textbf{99.14}\\ 97.41\\ 98.28\\ 47.41\\98.28\end{tabular}}&
\multicolumn{1}{l|}{\begin{tabular}[c]{@{}c@{}}\textbf{64.25}\\ 49.83\\ 50.08\\50.00\\ 51.42\end{tabular}}\\\cline{2-12}
\multicolumn{1}{|l|}{}&\multicolumn{1}{l|}{9}&
\multicolumn{1}{l|}{\begin{tabular}[c]{@{}c@{}}38.83\\ 44.15\\ \textbf{49.47}\\ 47.87\\ 45.75\end{tabular}}&
\multicolumn{1}{l|}{\begin{tabular}[c]{@{}c@{}}32.13\\ 35.74\\ 32.93\\ \textbf{66.27}\\35.94\end{tabular}}&
\multicolumn{1}{l|}{\begin{tabular}[c]{@{}c@{}}42.11\\ 30.06\\6.199\\\textbf{57.08}\\ 30.18\end{tabular}}&
\multicolumn{1}{l|}{\begin{tabular}[c]{@{}c@{}}38.32\\ 28.97\\ \textbf{59.81}\\\textbf{59.81}\\ 32.71\end{tabular}}&
\multicolumn{1}{l|}{\begin{tabular}[c]{@{}c@{}}41.18\\ \textbf{52.94}\\ 42.65\\ 41.18\\\textbf{52.94}\end{tabular}}&
\multicolumn{1}{l|}{\begin{tabular}[c]{@{}c@{}}52.02\\ 58.27\\ 46.42\\52.76\\ \textbf{58.64}\end{tabular}}&
\multicolumn{1}{l|}{\begin{tabular}[c]{@{}c@{}}\textbf{55.00}\\ 41.57\\43.02\\ 50.50\\ 41.27\end{tabular}}&
\multicolumn{1}{l|}{\begin{tabular}[c]{@{}c@{}}\textbf{98.33}\\ \textbf{98.33}\\ 50.00\\50.00\\ \textbf{98.33}\end{tabular}}&
\multicolumn{1}{l|}{}&
\multicolumn{1}{l|}{\begin{tabular}[c]{@{}c@{}}\textbf{61.42}\\ 53.58\\ 44.75\\50.00\\ 56.58\end{tabular}}\\\cline{2-12}
\multicolumn{1}{|l|}{}&
\multicolumn{1}{l|}{10}&
\multicolumn{1}{l|}{\begin{tabular}[c]{@{}c@{}}52.66\\ 88.30\\ 51.60\\54.79\\\textbf{94.15}\end{tabular}}&
\multicolumn{1}{l|}{\begin{tabular}[c]{@{}c@{}}\textbf{86.14}\\ 76.10\\ 66.87\\ 66.27\\ 74.50\end{tabular}}&
\multicolumn{1}{l|}{\begin{tabular}[c]{@{}c@{}}73.10\\ 85.03\\ 51.70\\ 59.88\\\textbf{86.90}\end{tabular}}&
\multicolumn{1}{l|}{\begin{tabular}[c]{@{}c@{}}46.73\\ \textbf{64.49}\\ 40.19\\ 40.19\\ 49.53\end{tabular}}&
\multicolumn{1}{l|}{\begin{tabular}[c]{@{}c@{}}76.47\\ 89.71\\ 58.82\\58.82\\ \textbf{98.53}\end{tabular}}&
\multicolumn{1}{l|}{\begin{tabular}[c]{@{}c@{}}86.49\\ 85.20\\ 49.45\\57.35\\ \textbf{94.58}\end{tabular}}&
\multicolumn{1}{l|}{\begin{tabular}[c]{@{}c@{}}68.27\\ 95.58\\ 48.74\\66.29\\ \textbf{96.57}\end{tabular}}&
\multicolumn{1}{l|}{\begin{tabular}[c]{@{}c@{}}\textbf{100.0}\\ 1.67\\ 50.00\\50.00\\ 83.33\end{tabular}}&
\multicolumn{1}{l|}{\begin{tabular}[c]{@{}c@{}}\textbf{100.0}\\ \textbf{100.0}\\0.00\\47.41\\ 99.14\end{tabular}}&
\multicolumn{1}{l|}{}\\\hline
\end{tabular}
\label{tab:all_res_gas}
\end{table}

\subsection{Experiments with Activity Recognition Data}
In this subsection, we present the experimental results obtained with other real data.
We used REALDISP activity recognition dataset~\cite{10.1145/2370216.2370437,s140609995}.
The REALDISP dataset is an open-access benchmark dataset for activity recognition systems that measures the 33 types of activities by 17 subjects by a set of motion capture sensors.
The original data in REALDISP dataset are time series obtained by 9 sensor units recorded at 50 Hz.
Each sensor unit records tri-directional acceleration, gyroscope, magnetic field measurements, as well as orientation estimated in quaternion format, and 13 values are observed for each sensor unit at each measurement; therefore, 117 values are obtained at each measurement in total.
A notable feature of the REALDISP dataset is that the effect of the sensor misplacement is considered.
In this experiment, we use the data of \textit{ideal} and \textit{self} scenarios.
In the \textit{ideal} scenario, data are obtained by the ideally placed sensors, and in the \textit{self} scenario, data are observed by the sensors placed by subjects themselves, which may be incorrectly placed.

We consider the domain adaptation problem where the source samples are the data of the ideal scenario and the target samples are the data of the self scenario.
Here, we assume that the shift between the source and target domains is caused by the sensor misplacement.
We use class L11: Waist bends forward and class L13: Waist bends (reach foot with opposite hand) for the classification and consider the binary classification problem.
Among the obtained sensor values, only the tri-directional accelerations of 9 sensors are used.
Furthermore, values of 5 out of 9 sensors are used to extract common features, and the features extracted from the rest of the sensors are used as the extra features.
To extract features from time series, a sliding window of 6 s with 2 s overlap is used, and we extracted the mean, standard deviation, maximum, minimum and mean crossing rate from each window.
Note that if no activity label is assigned in the window, the window is not used as training or test data.
Since no activity data is available for the self scenario of subjects 6 and 13; therefore, the self data of the rest of the 15 subjects are used as the target data.

We used data of ideal scenarios of all of the subjects as source samples and the data of self scenario of each subject as target samples.
Therefore, we consider the domain adaptation problem from 1 source domain to the 15 target domains.
Table \ref{tab:activity_recgnition} shows the accuracy in the target domain for each domain adaptation problem.
As we can see from the table, the accuracy of the prediction is improved by OT using extra features.
We also can see the effect of negative transfer for some target domains.

We conjecture that the success of the proposed method depends on the selection of the sensors to calculate the extra features.
When the misplaced sensors are selected to calculate the extra features, the distributions of $\bb x^e$ differ in the source and target domains.
In this case, the joint distribution of $(\bb x^c, y)$ is matched by the OT using the pseudo-labels, the distribution of $\bb x^e$ is changed due to the sensor misplacement, and the assumption Eq.~\eqref{eq:assumption_conditional} does not hold.
Therefore, in this case, the accuracy in the target domain can deteriorate.
In addition, the sensors that generate extra features are not necessarily informative for classification in general.
Therefore, accuracy of the proposed method may not improve or even decrease by considering

\begin{table}[ht]
    \centering
        \caption{Experimental results with REALDISP dataset}
    \begin{tabular}{P{8ex}cP{10ex}P{8ex}P{6ex}c|P{8ex}}
        \hline
        domains & Baseline & JDOT\ \ no extra& CCA & DSFT & Proposed & JDOT ideal\\
        \hline
        1 & 57.89 & 5.263 & \textbf{73.68} & 44.74 & 5.263 & 63.16\\
        2 & \textbf{69.23} & 61.54 & 61.54 & 46.15 & 61.54 & 87.18\\
        3 & 48.72 & 84.62 & 41.03 & 48.72 & \textbf{89.74} & 17.95\\
        4 & 86.84 & 86.84 & 50.00 & 47.37 & 86.84 & 97.37\\
        5 & 86.84 & \textbf{92.11} & 76.32 & 42.11 & 86.84 & 18.42\\
        6 & 88.57 & 88.57 & 62.86 & 48.57 & \textbf{100.0} & 100.0\\
        7 & 58.33 & \textbf{97.22} & 10.00 & 47.22 & \textbf{97.22} & 97.22\\
        8 & 47.37 & 44.74 & 81.58 & 47.37 & \textbf{97.37} & 97.37\\
        9 & \textbf{97.22} & 75.00 & 36.11 & 47.22 & 2.778 & 97.22\\
        10 & 93.33 & 51.11 & 51.11 & 48.89 & \textbf{95.56} & 95.56\\
        11 & 48.89 & 73.33 & 48.89 & 48.89 & \textbf{91.11} & 91.11\\
        12 & 72.41 & \textbf{75.86} & 51.72 & 72.41 & \textbf{75.86} & 75.86\\
        13 & \textbf{86.96} & 82.61 & 69.57 & 65.22 & 82.61 & 82.61\\
        14 & 93.18 & \textbf{97.73} & 59.09 & 50.00 & \textbf{97.73} & 97.73\\
        15 & \textbf{95.12} & 90.24 & 43.90 & 48.78 & 85.37 & 85.37\\
         \hline
    \end{tabular}
    \label{tab:activity_recgnition}
\end{table}
\clearpage

\bibliographystyle{apalike}
\bibliography{ref}
\end{document}